\documentclass{article} %
\usepackage[preprint]{conference}

\usepackage{microtype}
\usepackage{hyperref}
\usepackage{url}
\usepackage{booktabs}
\usepackage{todonotes}
\usepackage{amsmath, amssymb, amsthm, xspace, mathtools}
\usepackage[capitalize]{cleveref}

\usepackage{algorithm}
\usepackage{algorithmicx}
\usepackage[noend]{algpseudocode} %
\usepackage{tikz}
\usetikzlibrary{shadows}
\usepackage{longtable}

\usepackage{lineno}

\theoremstyle{plain}
\newtheorem{theorem}{Theorem}[section]

\newtheorem{lemma}[theorem]{Lemma}

\theoremstyle{definition}
\newtheorem{definition}[theorem]{Definition}

\theoremstyle{remark}

\newtheorem{problem}[theorem]{Problem}

\providecommand{\email}[1]{\href{mailto:#1}{\nolinkurl{#1}\xspace}}

\newcommand{\eps}{\ensuremath{\varepsilon}\xspace}
\newcommand{\eqdef}{:=}

\newcommand{\accept}{\textsf{accept}\xspace}

\newcommand{\reject}{\textsf{reject}\xspace}

\newcommand{\domain}{\ensuremath{\Omega}\xspace} %
\newcommand{\setOfSuchThat}[2]{ \left\{\; #1 \;\colon\; #2\; \right\} } 			%
\newcommand{\dtv}{\operatorname{d}_{\rm TV}}

\newcommand{\totalvardistrestr}[3][]{{\dtv^{#1}\mleft({#2, #3}\mright)}}
\newcommand{\totalvardist}[2]{\totalvardistrestr[]{#1}{#2}}

\newcommand\restr[2]{{%
  \left.\kern-\nulldelimiterspace %
  #1 %
  \vphantom{\big|} %
  \right|_{#2} %
  }}

\newcommand{\abs}[1]{\left\lvert #1 \right\rvert}

\newcommand{\clg}[1]{\left\lceil #1 \right\rceil}
\newcommand{\flr}[1]{\left\lfloor #1 \right\rfloor}

\newcommand{\pdfsamp}{dual\xspace}
\newcommand{\cdfsamp}{cumulative dual\xspace}
\newcommand{\Pdfsamp}{\expandafter\capitalisewords\expandafter{\pdfsamp}}
\newcommand{\Cdfsamp}{\expandafter\capitalisewords\expandafter{\cdfsamp}}

\makeatletter

\newcommand{\Rom}[1]{\expandafter\@slowromancap\romannumeral #1@}

\makeatother

\algnewcommand{\LineComment}[1]{\Statex \(\triangleright\) #1}

\newcommand{\p}{\mathbf{p}}
\newcommand{\q}{\mathbf{q}}

\newcommand{\cP}{\mathcal{P}}

\newcommand{\ns}{{\textcolor{red}{n}}} %
\newcommand{\ab}{\textcolor{cyan!80!black}{k}} %

\newcommand{\dst}{{\textcolor{ForestGreen!80!black}{\eps}}}
\newcommand{\errprob}{{\textcolor{brown}{\delta}}}

\newcommand{\cnt}{\operatorname{count}}
\newcommand{\worddict}{\Sigma_{w}}

\newcommand{\samp}{\ensuremath{\mathsf{SAMP}}\xspace}
\newcommand{\eval}{\ensuremath{\mathsf{EVAL}}\xspace}

\newcommand{\evalplus}{\ensuremath{\mathsf{EVAL^+}}\xspace}
\newcommand{\evalpp}{\ensuremath{\mathsf{TextEVAL^{+}}}\xspace}
\newcommand{\ideal}{\ensuremath{\mathcal{P}^*}}
\newcommand{\undertest}{\ensuremath{\mathcal{P}}}

\newcommand{\eos}{\ensuremath{\mathtt{EOS}}\xspace}

\newcommand{\tok}{\mathsf{tok}}

\newcommand{\enc}{\mathsf{enc}}
\newcommand{\dec}{\mathsf{dec}}
\newcommand{\tokdict}{\Sigma_{\mathsf{tok}}}
\newcommand{\cLtok}{\cL_{\tok}}
\newcommand{\Thresh}{\mathsf{Thresh}}

\newcommand{\bucketing}{\mathsf{Bucketing}}

\newcommand{\inbucket}{\mathsf{Local}}
\newcommand{\dkw}{\mathsf{Global}}
\newcommand{\cond}[2]{{#1}_{|{#2}}}

\newcommand{\stability}{\textsc{stable-code}}
\newcommand{\deepseek}{\textsc{deepseek-coder}}
\newcommand{\gemma}{\textsc{codegemma}}
\newcommand{\gpt}{\textsc{GPT2}}

\newcommand{\cS}{S}
\newcommand{\cT}{T}
\newcommand{\cL}{\mathcal{L}}

\newcommand{\cD}{\mathcal{D}}

\newcommand{\tOh}{\widetilde{{\mathcal O}}}

\newcommand{\getcoll}{\mathsf{getCollision}}
\newcommand{\alg}{\mathsf{Anubis}}
\newcommand{\base}{\mathsf{detectGPT}}

\newcommand{\bucket}{\Delta}
\newcommand{\model}{\ensuremath{\mathcal{L}^*}}
\newcommand{\hi}{\ensuremath{\mathsf{UB}}\xspace}
\newcommand{\lo}{\ensuremath{\mathsf{LB}}\xspace}
\newcommand{\bw}{\tau_{\Delta_0}}

\definecolor{darkblue}{rgb}{0, 0, 0.5}
\hypersetup{colorlinks=true, citecolor=darkblue, linkcolor=darkblue, urlcolor=darkblue}
\title{Zero-Shot Attribution for Large Language Models: \\ A Distribution Testing Approach\thanks{The accompanying tool, available open source, can be found at \protect\url{https://github.com/uddaloksarkar/anubis}.}}

\usepackage{authblk}

\author[1]{\textbf{Clément L. Canonne}}
\author[2]{\textbf{Yash Pote}}
\author[3]{\textbf{Uddalok Sarkar}}

\affil[1]{University of Sydney}
\affil[2]{National University of Singapore}
\affil[3]{Indian Statistical Institute}

\renewcommand{\paragraph}[1]{\noindent\textbf{#1}.}

\begin{document}

\ifcolmsubmission
\linenumbers
\fi

\maketitle

\begin{abstract}
A growing fraction of all code is sampled from Large Language Models (LLMs). We investigate the problem of attributing code generated by language models using hypothesis testing to leverage established techniques and guarantees. Given a set of samples $S$ and a suspect model $\mathcal{L}^*$, our goal is to assess the likelihood of $S$ originating from $\mathcal{L}^*$. Due to the curse of dimensionality, this is intractable when only samples from the LLM are given: to circumvent this, we use both samples and density estimates from the LLM, a form of access commonly available.

We introduce $\mathsf{Anubis}$, a zero-shot attribution tool that frames attribution as a distribution testing problem. Our experiments on a benchmark of code samples show that $\mathsf{Anubis}$ achieves high AUROC scores ( $\ge0.9$) when distinguishing between LLMs like DeepSeek-Coder, CodeGemma, and Stable-Code using only $\approx 2000$ samples. 
\end{abstract}

\section{Introduction}\label{sec:introduction}

Recent developments in large language models (LLMs) have led to a surge in machine-generated content across several domains, raising critical questions about attribution and accountability. As organizations increasingly deploy multiple LLMs for distinct use cases, distinguishing between outputs of different models has become essential for verifying compliance, tracking misuse, and understanding model-specific behaviors. This challenge is compounded by the quick evolution of LLMs, where new architectures and fine-tuned variants often produce text with overlapping statistical signatures.

A number of approaches have been proposed to solve the above challenge~\citep{wu2025survey}. However, long-term solutions remain elusive, as for most detectors, model updates and adversarial attacks are able to circumvent attribution. A fundamental question persists: \textit{What are the fundamental statistical limits on the performance of tests?}

In this work, we re-frame attribution as a \emph{distribution testing} problem. To connect the task of attribution to distribution testing, we think of LLMs as distributions over all possible token sequences.  Given a dataset, our goal reduces to determine whether the samples originate from a reference distribution $\cD^*$ (e.g., DeepSeek-Coder) or an alternative $\cD$ (e.g., Stable-code).

This abstraction allows us to leverage theoretical tools from distribution testing while addressing two practical challenges: 
(1) \textit{Dimensionality}: Direct comparison of high-dimensional  distributions (with support size $\approx (5\times 10^4)^{d}$), where $5\times 10^4$ is the size of the token set, and $d$ is the sequence length, is infeasible with standard i.i.d. sample bounds~\citep{VV:11:stoc}.
(2) \textit{Robustness}: LLM generated code undergoes post-processing (e.g., testing, formatting, dead code removal), violating idealized distributional assumptions.

To address this issue, we adopt a framework from \emph{oracle-enhanced} distribution testing, a sub-field of testing that focuses on the use of more powerful \textit{oracles} beyond sampling to obtain more data-efficient statistical tests. For instance, an \emph{evaluation oracle} $\eval(\cD,x)$ provides, for a given data point $x$, the probability to see $x$ occur under distribution $\cD$. Combining this with the ability to sample (modeled as a \emph{sampling oracle} $\samp(\cD)$) can, in theory, lead to much more data-efficient algorithms, bypassing the curse of dimensionality~\citep{CRS:14,OnakS18}. 

This paper can thus be seen as porting these theoretical results from the distribution testing literature to the practical, high-dimensional setting of attribution to assess whether these data efficiency gains translate in the real world. This approach presents, on the one hand, some very appealing features: it is completely agnostic to post-processing and does not require any ad hoc method such as watermarking. On the other hand, it does present some significant challenges: one may wonder whether this domain-agnostic, abstracted approach may have to pay a stark efficiency price due to its very generality. Moreover, as our paper demonstrates, a significant amount of work is required to translate the theoretical guarantees obtained in the distribution testing literature (where constant factors are typically ignored, and simplifying modeling assumptions are common) to our practical setting, where the constant factors do have a significant impact on the result, and robustness to deviations from an idealized model is a necessity. 

\begin{figure*}
    \centering

\tikzset{every picture/.style={line width=0.75pt}} %

\begin{tikzpicture}[x=0.6pt,y=0.6pt,yscale=-1,xscale=1]
\draw  [fill={rgb, 255:red, 173; green, 217; blue, 244 }  ,fill opacity=0.12 ] (651.9,428.67) -- (651.9,492.3) -- (466.83,492.3) -- (466.83,507.87) -- (472.7,507.87) -- (463.9,518.07) -- (455.1,507.87) -- (460.97,507.87) -- (460.97,492.3) -- (275.9,492.3) -- (275.9,428.67) -- cycle ;
\draw  [fill={rgb, 255:red, 134; green, 186; blue, 233 }  ,fill opacity=0.14 ] (147.9,230.83) .. controls (147.9,194.07) and (177.7,164.27) .. (214.46,164.27) -- (726.34,164.27) .. controls (763.1,164.27) and (792.9,194.07) .. (792.9,230.83) -- (792.9,430.51) .. controls (792.9,467.28) and (763.1,497.07) .. (726.34,497.07) -- (214.46,497.07) .. controls (177.7,497.07) and (147.9,467.28) .. (147.9,430.51) -- cycle ;
\draw  [color={rgb, 255:red, 70; green, 140; blue, 152 }  ,draw opacity=0.52 ][fill={rgb, 255:red, 70; green, 140; blue, 152 }  ,fill opacity=0.26 ] (185,387.02) -- (403.9,387.02) -- (403.9,408.27) -- (185,408.27) -- cycle ;
\draw  [color={rgb, 255:red, 155; green, 155; blue, 155 }  ,draw opacity=0.48 ] (215.57,325.1) -- (215.57,382.2) .. controls (215.57,384.86) and (208.38,387.02) .. (199.5,387.02) .. controls (190.62,387.02) and (183.43,384.86) .. (183.43,382.2) -- (183.43,325.1) .. controls (183.43,322.43) and (190.62,320.27) .. (199.5,320.27) .. controls (208.38,320.27) and (215.57,322.43) .. (215.57,325.1) .. controls (215.57,327.76) and (208.38,329.92) .. (199.5,329.92) .. controls (190.62,329.92) and (183.43,327.76) .. (183.43,325.1) ;
\draw  [color={rgb, 255:red, 155; green, 155; blue, 155 }  ,draw opacity=0.48 ] (247.71,325.1) -- (247.71,382.2) .. controls (247.71,384.86) and (240.52,387.02) .. (231.64,387.02) .. controls (222.77,387.02) and (215.57,384.86) .. (215.57,382.2) -- (215.57,325.1) .. controls (215.57,322.43) and (222.77,320.27) .. (231.64,320.27) .. controls (240.52,320.27) and (247.71,322.43) .. (247.71,325.1) .. controls (247.71,327.76) and (240.52,329.92) .. (231.64,329.92) .. controls (222.77,329.92) and (215.57,327.76) .. (215.57,325.1) ;
\draw  [color={rgb, 255:red, 155; green, 155; blue, 155 }  ,draw opacity=0.48 ] (279.86,325.1) -- (279.86,382.2) .. controls (279.86,384.86) and (272.66,387.02) .. (263.79,387.02) .. controls (254.91,387.02) and (247.71,384.86) .. (247.71,382.2) -- (247.71,325.1) .. controls (247.71,322.43) and (254.91,320.27) .. (263.79,320.27) .. controls (272.66,320.27) and (279.86,322.43) .. (279.86,325.1) .. controls (279.86,327.76) and (272.66,329.92) .. (263.79,329.92) .. controls (254.91,329.92) and (247.71,327.76) .. (247.71,325.1) ;
\draw  [color={rgb, 255:red, 155; green, 155; blue, 155 }  ,draw opacity=0.48 ] (312,325.1) -- (312,382.2) .. controls (312,384.86) and (304.8,387.02) .. (295.93,387.02) .. controls (287.05,387.02) and (279.86,384.86) .. (279.86,382.2) -- (279.86,325.1) .. controls (279.86,322.43) and (287.05,320.27) .. (295.93,320.27) .. controls (304.8,320.27) and (312,322.43) .. (312,325.1) .. controls (312,327.76) and (304.8,329.92) .. (295.93,329.92) .. controls (287.05,329.92) and (279.86,327.76) .. (279.86,325.1) ;
\draw  [color={rgb, 255:red, 155; green, 155; blue, 155 }  ,draw opacity=0.48 ] (344.14,325.1) -- (344.14,382.2) .. controls (344.14,384.86) and (336.95,387.02) .. (328.07,387.02) .. controls (319.2,387.02) and (312,384.86) .. (312,382.2) -- (312,325.1) .. controls (312,322.43) and (319.2,320.27) .. (328.07,320.27) .. controls (336.95,320.27) and (344.14,322.43) .. (344.14,325.1) .. controls (344.14,327.76) and (336.95,329.92) .. (328.07,329.92) .. controls (319.2,329.92) and (312,327.76) .. (312,325.1) ;
\draw  [color={rgb, 255:red, 155; green, 155; blue, 155 }  ,draw opacity=0.48 ] (376.29,325.1) -- (376.29,382.2) .. controls (376.29,384.86) and (369.09,387.02) .. (360.21,387.02) .. controls (351.34,387.02) and (344.14,384.86) .. (344.14,382.2) -- (344.14,325.1) .. controls (344.14,322.43) and (351.34,320.27) .. (360.21,320.27) .. controls (369.09,320.27) and (376.29,322.43) .. (376.29,325.1) .. controls (376.29,327.76) and (369.09,329.92) .. (360.21,329.92) .. controls (351.34,329.92) and (344.14,327.76) .. (344.14,325.1) ;
\draw  [color={rgb, 255:red, 155; green, 155; blue, 155 }  ,draw opacity=0.48 ] (408.43,325.1) -- (408.43,382.2) .. controls (408.43,384.86) and (401.23,387.02) .. (392.36,387.02) .. controls (383.48,387.02) and (376.29,384.86) .. (376.29,382.2) -- (376.29,325.1) .. controls (376.29,322.43) and (383.48,320.27) .. (392.36,320.27) .. controls (401.23,320.27) and (408.43,322.43) .. (408.43,325.1) .. controls (408.43,327.76) and (401.23,329.92) .. (392.36,329.92) .. controls (383.48,329.92) and (376.29,327.76) .. (376.29,325.1) ;
\draw  [color={rgb, 255:red, 0; green, 0; blue, 0 }  ,draw opacity=1 ][fill={rgb, 255:red, 233; green, 191; blue, 134 }  ,fill opacity=1 ] (215.57,366.99) -- (215.57,383.49) .. controls (215.57,385.44) and (208.38,387.02) .. (199.5,387.02) .. controls (190.62,387.02) and (183.43,385.44) .. (183.43,383.49) -- (183.43,366.99) .. controls (183.43,365.04) and (190.62,363.46) .. (199.5,363.46) .. controls (208.38,363.46) and (215.57,365.04) .. (215.57,366.99) .. controls (215.57,368.95) and (208.38,370.53) .. (199.5,370.53) .. controls (190.62,370.53) and (183.43,368.95) .. (183.43,366.99) ;
\draw  [color={rgb, 255:red, 0; green, 0; blue, 0 }  ,draw opacity=1 ][fill={rgb, 255:red, 233; green, 191; blue, 134 }  ,fill opacity=1 ] (247.71,355.83) -- (247.71,382.2) .. controls (247.71,384.86) and (240.52,387.02) .. (231.64,387.02) .. controls (222.77,387.02) and (215.57,384.86) .. (215.57,382.2) -- (215.57,355.83) .. controls (215.57,353.17) and (222.77,351.01) .. (231.64,351.01) .. controls (240.52,351.01) and (247.71,353.17) .. (247.71,355.83) .. controls (247.71,358.5) and (240.52,360.66) .. (231.64,360.66) .. controls (222.77,360.66) and (215.57,358.5) .. (215.57,355.83) ;
\draw  [color={rgb, 255:red, 0; green, 0; blue, 0 }  ,draw opacity=1 ][fill={rgb, 255:red, 233; green, 191; blue, 134 }  ,fill opacity=1 ] (279.86,361.7) -- (279.86,382.55) .. controls (279.86,385.02) and (272.66,387.02) .. (263.79,387.02) .. controls (254.91,387.02) and (247.71,385.02) .. (247.71,382.55) -- (247.71,361.7) .. controls (247.71,359.24) and (254.91,357.24) .. (263.79,357.24) .. controls (272.66,357.24) and (279.86,359.24) .. (279.86,361.7) .. controls (279.86,364.17) and (272.66,366.17) .. (263.79,366.17) .. controls (254.91,366.17) and (247.71,364.17) .. (247.71,361.7) ;
\draw  [color={rgb, 255:red, 0; green, 0; blue, 0 }  ,draw opacity=1 ][fill={rgb, 255:red, 233; green, 191; blue, 134 }  ,fill opacity=1 ] (312,338.5) -- (312,382.2) .. controls (312,384.86) and (304.8,387.02) .. (295.93,387.02) .. controls (287.05,387.02) and (279.86,384.86) .. (279.86,382.2) -- (279.86,338.5) .. controls (279.86,335.83) and (287.05,333.67) .. (295.93,333.67) .. controls (304.8,333.67) and (312,335.83) .. (312,338.5) .. controls (312,341.16) and (304.8,343.32) .. (295.93,343.32) .. controls (287.05,343.32) and (279.86,341.16) .. (279.86,338.5) ;
\draw  [color={rgb, 255:red, 0; green, 0; blue, 0 }  ,draw opacity=1 ][fill={rgb, 255:red, 233; green, 191; blue, 134 }  ,fill opacity=1 ] (344.14,335.38) -- (344.14,382.2) .. controls (344.14,384.86) and (336.95,387.02) .. (328.07,387.02) .. controls (319.2,387.02) and (312,384.86) .. (312,382.2) -- (312,335.38) .. controls (312,332.72) and (319.2,330.56) .. (328.07,330.56) .. controls (336.95,330.56) and (344.14,332.72) .. (344.14,335.38) .. controls (344.14,338.05) and (336.95,340.21) .. (328.07,340.21) .. controls (319.2,340.21) and (312,338.05) .. (312,335.38) ;
\draw  [color={rgb, 255:red, 0; green, 0; blue, 0 }  ,draw opacity=1 ][fill={rgb, 255:red, 233; green, 191; blue, 134 }  ,fill opacity=1 ] (376.29,347.83) -- (376.29,382.2) .. controls (376.29,384.86) and (369.09,387.02) .. (360.21,387.02) .. controls (351.34,387.02) and (344.14,384.86) .. (344.14,382.2) -- (344.14,347.83) .. controls (344.14,345.17) and (351.34,343.01) .. (360.21,343.01) .. controls (369.09,343.01) and (376.29,345.17) .. (376.29,347.83) .. controls (376.29,350.49) and (369.09,352.65) .. (360.21,352.65) .. controls (351.34,352.65) and (344.14,350.49) .. (344.14,347.83) ;
\draw  [color={rgb, 255:red, 0; green, 0; blue, 0 }  ,draw opacity=1 ][fill={rgb, 255:red, 233; green, 191; blue, 134 }  ,fill opacity=1 ] (408.43,340.27) -- (408.43,382.2) .. controls (408.43,384.86) and (401.23,387.02) .. (392.36,387.02) .. controls (383.48,387.02) and (376.29,384.86) .. (376.29,382.2) -- (376.29,340.27) .. controls (376.29,337.61) and (383.48,335.45) .. (392.36,335.45) .. controls (401.23,335.45) and (408.43,337.61) .. (408.43,340.27) .. controls (408.43,342.94) and (401.23,345.1) .. (392.36,345.1) .. controls (383.48,345.1) and (376.29,342.94) .. (376.29,340.27) ;
\draw  [color={rgb, 255:red, 155; green, 155; blue, 155 }  ,draw opacity=0.48 ] (567.57,326.35) -- (567.57,383.45) .. controls (567.57,386.12) and (560.38,388.27) .. (551.5,388.27) .. controls (542.62,388.27) and (535.43,386.12) .. (535.43,383.45) -- (535.43,326.35) .. controls (535.43,323.69) and (542.62,321.53) .. (551.5,321.53) .. controls (560.38,321.53) and (567.57,323.69) .. (567.57,326.35) .. controls (567.57,329.01) and (560.38,331.17) .. (551.5,331.17) .. controls (542.62,331.17) and (535.43,329.01) .. (535.43,326.35) ;
\draw  [color={rgb, 255:red, 155; green, 155; blue, 155 }  ,draw opacity=0.48 ] (599.71,326.35) -- (599.71,383.45) .. controls (599.71,386.12) and (592.52,388.27) .. (583.64,388.27) .. controls (574.77,388.27) and (567.57,386.12) .. (567.57,383.45) -- (567.57,326.35) .. controls (567.57,323.69) and (574.77,321.53) .. (583.64,321.53) .. controls (592.52,321.53) and (599.71,323.69) .. (599.71,326.35) .. controls (599.71,329.01) and (592.52,331.17) .. (583.64,331.17) .. controls (574.77,331.17) and (567.57,329.01) .. (567.57,326.35) ;
\draw  [color={rgb, 255:red, 155; green, 155; blue, 155 }  ,draw opacity=0.48 ] (631.86,326.35) -- (631.86,383.45) .. controls (631.86,386.12) and (624.66,388.27) .. (615.79,388.27) .. controls (606.91,388.27) and (599.71,386.12) .. (599.71,383.45) -- (599.71,326.35) .. controls (599.71,323.69) and (606.91,321.53) .. (615.79,321.53) .. controls (624.66,321.53) and (631.86,323.69) .. (631.86,326.35) .. controls (631.86,329.01) and (624.66,331.17) .. (615.79,331.17) .. controls (606.91,331.17) and (599.71,329.01) .. (599.71,326.35) ;
\draw  [color={rgb, 255:red, 155; green, 155; blue, 155 }  ,draw opacity=0.48 ] (664,326.35) -- (664,383.45) .. controls (664,386.12) and (656.8,388.27) .. (647.93,388.27) .. controls (639.05,388.27) and (631.86,386.12) .. (631.86,383.45) -- (631.86,326.35) .. controls (631.86,323.69) and (639.05,321.53) .. (647.93,321.53) .. controls (656.8,321.53) and (664,323.69) .. (664,326.35) .. controls (664,329.01) and (656.8,331.17) .. (647.93,331.17) .. controls (639.05,331.17) and (631.86,329.01) .. (631.86,326.35) ;
\draw  [color={rgb, 255:red, 155; green, 155; blue, 155 }  ,draw opacity=0.48 ] (696.14,326.35) -- (696.14,383.45) .. controls (696.14,386.12) and (688.95,388.27) .. (680.07,388.27) .. controls (671.2,388.27) and (664,386.12) .. (664,383.45) -- (664,326.35) .. controls (664,323.69) and (671.2,321.53) .. (680.07,321.53) .. controls (688.95,321.53) and (696.14,323.69) .. (696.14,326.35) .. controls (696.14,329.01) and (688.95,331.17) .. (680.07,331.17) .. controls (671.2,331.17) and (664,329.01) .. (664,326.35) ;
\draw  [color={rgb, 255:red, 155; green, 155; blue, 155 }  ,draw opacity=0.48 ] (728.29,326.35) -- (728.29,383.45) .. controls (728.29,386.12) and (721.09,388.27) .. (712.21,388.27) .. controls (703.34,388.27) and (696.14,386.12) .. (696.14,383.45) -- (696.14,326.35) .. controls (696.14,323.69) and (703.34,321.53) .. (712.21,321.53) .. controls (721.09,321.53) and (728.29,323.69) .. (728.29,326.35) .. controls (728.29,329.01) and (721.09,331.17) .. (712.21,331.17) .. controls (703.34,331.17) and (696.14,329.01) .. (696.14,326.35) ;
\draw  [color={rgb, 255:red, 155; green, 155; blue, 155 }  ,draw opacity=0.48 ] (760.43,326.35) -- (760.43,383.45) .. controls (760.43,386.12) and (753.23,388.27) .. (744.36,388.27) .. controls (735.48,388.27) and (728.29,386.12) .. (728.29,383.45) -- (728.29,326.35) .. controls (728.29,323.69) and (735.48,321.53) .. (744.36,321.53) .. controls (753.23,321.53) and (760.43,323.69) .. (760.43,326.35) .. controls (760.43,329.01) and (753.23,331.17) .. (744.36,331.17) .. controls (735.48,331.17) and (728.29,329.01) .. (728.29,326.35) ;
\draw  [color={rgb, 255:red, 0; green, 0; blue, 0 }  ,draw opacity=1 ][fill={rgb, 255:red, 173; green, 217; blue, 244 }  ,fill opacity=1 ] (567.57,368.25) -- (567.57,384.74) .. controls (567.57,386.69) and (560.38,388.27) .. (551.5,388.27) .. controls (542.62,388.27) and (535.43,386.69) .. (535.43,384.74) -- (535.43,368.25) .. controls (535.43,366.3) and (542.62,364.71) .. (551.5,364.71) .. controls (560.38,364.71) and (567.57,366.3) .. (567.57,368.25) .. controls (567.57,370.2) and (560.38,371.78) .. (551.5,371.78) .. controls (542.62,371.78) and (535.43,370.2) .. (535.43,368.25) ;
\draw  [color={rgb, 255:red, 0; green, 0; blue, 0 }  ,draw opacity=1 ][fill={rgb, 255:red, 173; green, 217; blue, 244 }  ,fill opacity=1 ] (599.71,357.09) -- (599.71,383.45) .. controls (599.71,386.12) and (592.52,388.27) .. (583.64,388.27) .. controls (574.77,388.27) and (567.57,386.12) .. (567.57,383.45) -- (567.57,357.09) .. controls (567.57,354.43) and (574.77,352.27) .. (583.64,352.27) .. controls (592.52,352.27) and (599.71,354.43) .. (599.71,357.09) .. controls (599.71,359.75) and (592.52,361.91) .. (583.64,361.91) .. controls (574.77,361.91) and (567.57,359.75) .. (567.57,357.09) ;
\draw  [color={rgb, 255:red, 0; green, 0; blue, 0 }  ,draw opacity=1 ][fill={rgb, 255:red, 173; green, 217; blue, 244 }  ,fill opacity=1 ] (631.86,344.39) -- (631.86,383.45) .. controls (631.86,386.12) and (624.66,388.27) .. (615.79,388.27) .. controls (606.91,388.27) and (599.71,386.12) .. (599.71,383.45) -- (599.71,344.39) .. controls (599.71,341.72) and (606.91,339.57) .. (615.79,339.57) .. controls (624.66,339.57) and (631.86,341.72) .. (631.86,344.39) .. controls (631.86,347.05) and (624.66,349.21) .. (615.79,349.21) .. controls (606.91,349.21) and (599.71,347.05) .. (599.71,344.39) ;
\draw  [color={rgb, 255:red, 0; green, 0; blue, 0 }  ,draw opacity=1 ][fill={rgb, 255:red, 173; green, 217; blue, 244 }  ,fill opacity=1 ] (664,357.72) -- (664,383.45) .. controls (664,386.12) and (656.8,388.27) .. (647.93,388.27) .. controls (639.05,388.27) and (631.86,386.12) .. (631.86,383.45) -- (631.86,357.72) .. controls (631.86,355.06) and (639.05,352.9) .. (647.93,352.9) .. controls (656.8,352.9) and (664,355.06) .. (664,357.72) .. controls (664,360.39) and (656.8,362.54) .. (647.93,362.54) .. controls (639.05,362.54) and (631.86,360.39) .. (631.86,357.72) ;
\draw  [color={rgb, 255:red, 0; green, 0; blue, 0 }  ,draw opacity=1 ][fill={rgb, 255:red, 173; green, 217; blue, 244 }  ,fill opacity=1 ] (696.14,361.61) -- (696.14,383.57) .. controls (696.14,386.17) and (688.95,388.27) .. (680.07,388.27) .. controls (671.2,388.27) and (664,386.17) .. (664,383.57) -- (664,361.61) .. controls (664,359.01) and (671.2,356.9) .. (680.07,356.9) .. controls (688.95,356.9) and (696.14,359.01) .. (696.14,361.61) .. controls (696.14,364.21) and (688.95,366.31) .. (680.07,366.31) .. controls (671.2,366.31) and (664,364.21) .. (664,361.61) ;
\draw  [color={rgb, 255:red, 0; green, 0; blue, 0 }  ,draw opacity=1 ][fill={rgb, 255:red, 173; green, 217; blue, 244 }  ,fill opacity=1 ] (728.29,341.72) -- (728.29,383.45) .. controls (728.29,386.12) and (721.09,388.27) .. (712.21,388.27) .. controls (703.34,388.27) and (696.14,386.12) .. (696.14,383.45) -- (696.14,341.72) .. controls (696.14,339.06) and (703.34,336.9) .. (712.21,336.9) .. controls (721.09,336.9) and (728.29,339.06) .. (728.29,341.72) .. controls (728.29,344.38) and (721.09,346.54) .. (712.21,346.54) .. controls (703.34,346.54) and (696.14,344.38) .. (696.14,341.72) ;
\draw  [color={rgb, 255:red, 0; green, 0; blue, 0 }  ,draw opacity=1 ][fill={rgb, 255:red, 173; green, 217; blue, 244 }  ,fill opacity=1 ] (760.43,341.53) -- (760.43,383.45) .. controls (760.43,386.12) and (753.23,388.27) .. (744.36,388.27) .. controls (735.48,388.27) and (728.29,386.12) .. (728.29,383.45) -- (728.29,341.53) .. controls (728.29,338.87) and (735.48,336.71) .. (744.36,336.71) .. controls (753.23,336.71) and (760.43,338.87) .. (760.43,341.53) .. controls (760.43,344.19) and (753.23,346.35) .. (744.36,346.35) .. controls (735.48,346.35) and (728.29,344.19) .. (728.29,341.53) ;
\draw (557.9,176.77) node  {\includegraphics[width=70pt,height=56pt]{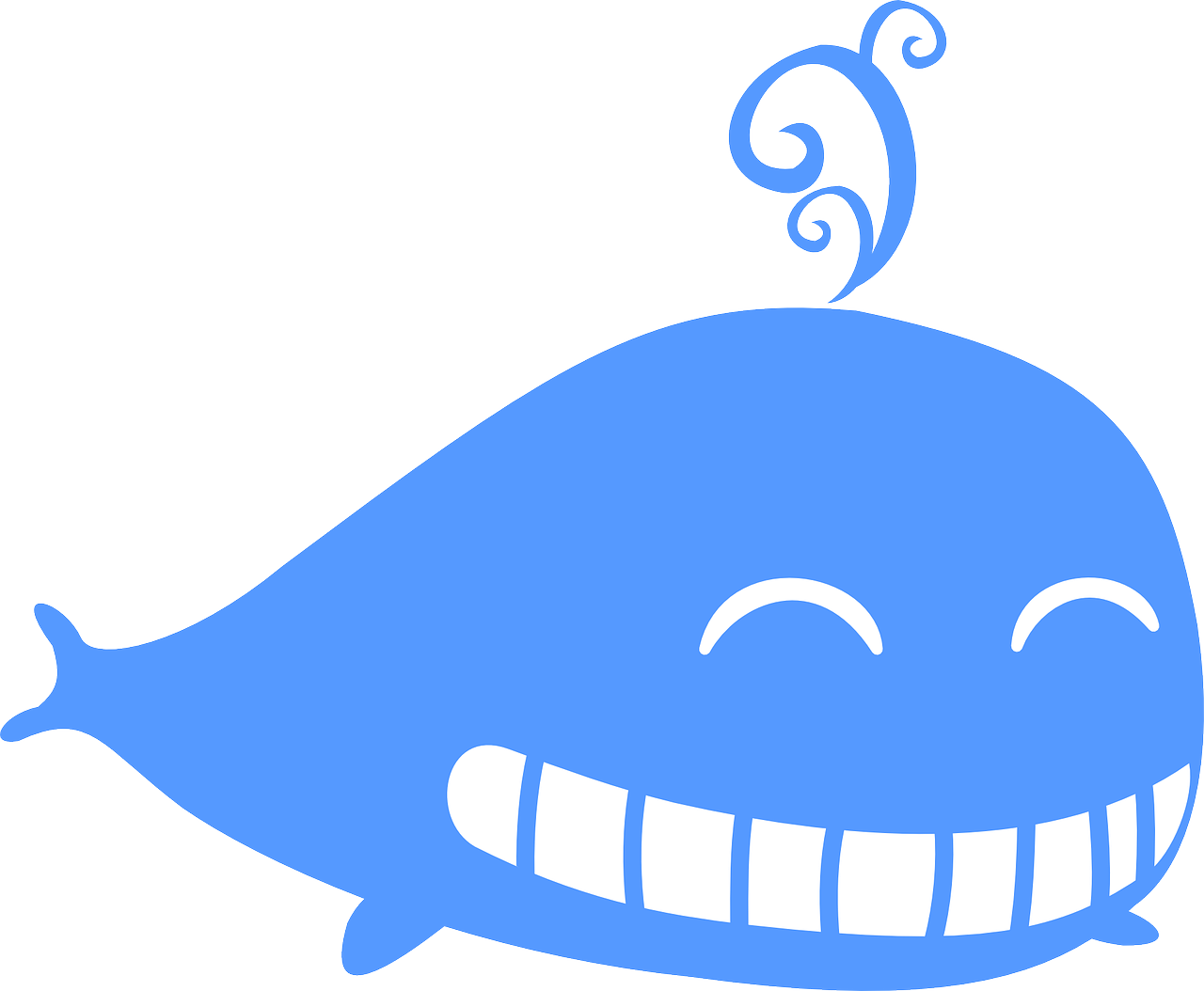}};
\draw  [fill={rgb, 255:red, 232; green, 232; blue, 215 }  ,fill opacity=1 ] (340.86,165.77) -- (340.86,216.04) .. controls (340.86,221.99) and (319.85,226.81) .. (293.93,226.81) .. controls (268.01,226.81) and (247,221.99) .. (247,216.04) -- (247,165.77) .. controls (247,159.82) and (268.01,155) .. (293.93,155) .. controls (319.85,155) and (340.86,159.82) .. (340.86,165.77) .. controls (340.86,171.72) and (319.85,176.54) .. (293.93,176.54) .. controls (268.01,176.54) and (247,171.72) .. (247,165.77) ;

\draw [color={rgb, 255:red, 97; green, 123; blue, 191 }  ,draw opacity=1 ][line width=1.5]    (504.9,209.27) .. controls (419.2,216.76) and (296.34,244.96) .. (231.81,273.38) ;
\draw [shift={(228.9,274.67)}, rotate = 335.63] [fill={rgb, 255:red, 97; green, 123; blue, 191 }  ,fill opacity=1 ][line width=0.08]  [draw opacity=0] (11.61,-5.58) -- (0,0) -- (11.61,5.58) -- cycle    ;
\draw [color={rgb, 255:red, 97; green, 123; blue, 191 }  ,draw opacity=1 ][line width=1.5]    (504.9,209.27) .. controls (432.9,225.21) and (408.82,247.41) .. (391.09,271.64) ;
\draw [shift={(388.9,274.67)}, rotate = 305.32] [fill={rgb, 255:red, 97; green, 123; blue, 191 }  ,fill opacity=1 ][line width=0.08]  [draw opacity=0] (11.61,-5.58) -- (0,0) -- (11.61,5.58) -- cycle    ;
\draw [color={rgb, 255:red, 71; green, 108; blue, 155 }  ,draw opacity=1 ][line width=1.5]    (606,229.33) .. controls (634.05,234.94) and (697.07,265.03) .. (714.95,277.09) ;
\draw [shift={(718,279.33)}, rotate = 219.81] [fill={rgb, 255:red, 71; green, 108; blue, 155 }  ,fill opacity=1 ][line width=0.08]  [draw opacity=0] (11.61,-5.58) -- (0,0) -- (11.61,5.58) -- cycle    ;
\draw [color={rgb, 255:red, 71; green, 108; blue, 155 }  ,draw opacity=1 ][line width=1.5]    (577,229.33) .. controls (577,240.74) and (577.77,253.68) .. (583.36,267.49) .. controls (583.66,268.24) and (583.98,268.98) .. (584.31,269.73) ;
\draw [shift={(586,273.33)}, rotate = 243.43] [fill={rgb, 255:red, 71; green, 108; blue, 155 }  ,fill opacity=1 ][line width=0.08]  [draw opacity=0] (11.61,-5.58) -- (0,0) -- (11.61,5.58) -- cycle    ;
\draw  [color={rgb, 255:red, 78; green, 165; blue, 217 }  ,draw opacity=0.56 ][fill={rgb, 255:red, 78; green, 165; blue, 217 }  ,fill opacity=0.56 ] (332.78,408.67) -- (332.78,441.14) -- (495.39,441.14) -- (495.39,429.73) -- (514,445.53) -- (495.39,461.33) -- (495.39,449.92) -- (324,449.92) -- (324,408.67) -- cycle ;
\draw  [color={rgb, 255:red, 78; green, 165; blue, 217 }  ,draw opacity=0.56 ][fill={rgb, 255:red, 78; green, 165; blue, 217 }  ,fill opacity=0.58 ] (675.71,411.33) -- (675.71,443.47) -- (583.44,443.47) -- (583.44,434.15) -- (564.67,447.61) -- (583.44,461.07) -- (583.44,451.75) -- (684,451.75) -- (684,411.33) -- cycle ;
\draw  [color={rgb, 255:red, 70; green, 140; blue, 152 }  ,draw opacity=0.53 ][fill={rgb, 255:red, 97; green, 123; blue, 191 }  ,fill opacity=1 ] (254.9,409.24) -- (254.9,469.96) -- (299.9,469.96) -- (299.9,458.07) -- (324,475.07) -- (299.9,492.07) -- (299.9,480.19) -- (244.67,480.19) -- (244.67,409.24) -- cycle ;
\draw  [color={rgb, 255:red, 70; green, 140; blue, 152 }  ,draw opacity=0.53 ][fill={rgb, 255:red, 97; green, 123; blue, 191 }  ,fill opacity=1 ] (713.9,410) -- (713.9,471.41) -- (396.64,471.41) -- (396.64,461.75) -- (377.9,476.41) -- (396.64,491.07) -- (396.64,481.41) -- (723.9,481.41) -- (723.9,410) -- cycle ;
\draw  [color={rgb, 255:red, 70; green, 140; blue, 152 }  ,draw opacity=0.26 ][fill={rgb, 255:red, 70; green, 140; blue, 152 }  ,fill opacity=0.26 ] (537,390.02) -- (755.9,390.02) -- (755.9,411.27) -- (537,411.27) -- cycle ;
\draw  [color={rgb, 255:red, 78; green, 165; blue, 217 }  ,draw opacity=0.56 ][fill={rgb, 255:red, 78; green, 165; blue, 217 }  ,fill opacity=0.57 ] (317.57,398.52) .. controls (317.57,392.72) and (322.27,388.02) .. (328.07,388.02) .. controls (333.87,388.02) and (338.57,392.72) .. (338.57,398.52) .. controls (338.57,404.32) and (333.87,409.02) .. (328.07,409.02) .. controls (322.27,409.02) and (317.57,404.32) .. (317.57,398.52) -- cycle ;
\draw  [color={rgb, 255:red, 78; green, 165; blue, 217 }  ,draw opacity=0.57 ][fill={rgb, 255:red, 78; green, 165; blue, 217 }  ,fill opacity=0.58 ] (669.57,401.77) .. controls (669.57,395.98) and (674.27,391.27) .. (680.07,391.27) .. controls (685.87,391.27) and (690.57,395.98) .. (690.57,401.77) .. controls (690.57,407.57) and (685.87,412.27) .. (680.07,412.27) .. controls (674.27,412.27) and (669.57,407.57) .. (669.57,401.77) -- cycle ;
\draw (473.09,334.59) node  {\includegraphics[width=50pt,height=50pt]{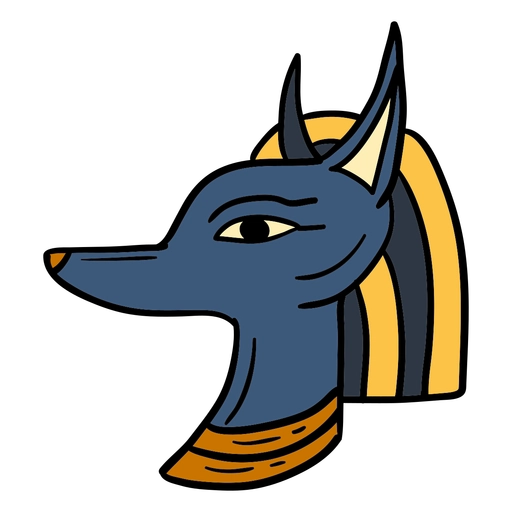}};
\draw (440,380) node [anchor=north west][inner sep=0.75pt]  [font=\Large]  {$\alg$};
\draw  [color={rgb, 255:red, 246; green, 203; blue, 118 }  ,draw opacity=1 ][fill={rgb, 255:red, 246; green, 203; blue, 118 }  ,fill opacity=1 ] (352.65,187.67) -- (472.82,187.67) -- (472.82,180.07) -- (496.9,191.37) -- (472.82,202.67) -- (472.82,195.07) -- (352.65,195.07) -- cycle ;\draw  [color={rgb, 255:red, 246; green, 203; blue, 118 }  ,draw opacity=1 ][fill={rgb, 255:red, 246; green, 203; blue, 118 }  ,fill opacity=1 ] (347,187.67) -- (348.13,187.67) -- (348.13,195.07) -- (347,195.07) -- cycle ;\draw  [color={rgb, 255:red, 246; green, 203; blue, 118 }  ,draw opacity=1 ][fill={rgb, 255:red, 246; green, 203; blue, 118 }  ,fill opacity=1 ] (349.26,187.67) -- (351.52,187.67) -- (351.52,195.07) -- (349.26,195.07) -- cycle ;
\draw  [color={rgb, 255:red, 162; green, 187; blue, 232 }  ,draw opacity=1 ][fill={rgb, 255:red, 162; green, 187; blue, 232 }  ,fill opacity=1 ] (677.9,197.39) .. controls (677.9,187.17) and (651.08,178.89) .. (618,178.89) -- (618,171.27) .. controls (651.08,171.27) and (677.9,179.56) .. (677.9,189.77) ;\draw  [color={rgb, 255:red, 162; green, 187; blue, 232 }  ,draw opacity=1 ][fill={rgb, 255:red, 162; green, 187; blue, 232 }  ,fill opacity=1 ] (677.9,189.77) .. controls (677.9,197.62) and (662.08,204.32) .. (639.76,207.01) -- (639.76,197.22) -- (618,212.07) -- (639.76,224.42) -- (639.76,214.62) .. controls (662.08,211.94) and (677.9,205.23) .. (677.9,197.39)(677.9,189.77) -- (677.9,197.39) ;
\draw  [color={rgb, 255:red, 162; green, 187; blue, 232 }  ,draw opacity=1 ][fill={rgb, 255:red, 162; green, 187; blue, 232 }  ,fill opacity=1 ] (185.86,296.72) -- (408.86,296.72) -- (408.86,314.72) -- (185.86,314.72) -- cycle ;

\draw  [color={rgb, 255:red, 162; green, 187; blue, 232 }  ,draw opacity=1 ][fill={rgb, 255:red, 162; green, 187; blue, 232 }  ,fill opacity=1 ] (535.86,297.72) -- (758.86,297.72) -- (758.86,315.72) -- (535.86,315.72) -- cycle ;

\draw  [color={rgb, 255:red, 246; green, 203; blue, 118 }  ,draw opacity=1 ][fill={rgb, 255:red, 246; green, 203; blue, 118 }  ,fill opacity=1 ] (237,217.07) -- (199.49,217.07) -- (199.49,259.07) -- (205.9,259.07) -- (196.45,271) -- (187,259.07) -- (193.41,259.07) -- (193.41,211) -- (237,211) -- cycle ;
\draw  [color={rgb, 255:red, 246; green, 203; blue, 118 }  ,draw opacity=1 ][fill={rgb, 255:red, 246; green, 203; blue, 118 }  ,fill opacity=1 ] (344.9,216.35) -- (351.78,216.35) -- (351.78,259.29) -- (344.9,259.29) -- (354.95,272) -- (365,259.29) -- (358.12,259.29) -- (358.12,210) -- (344.9,210) -- cycle ;

\draw (188,275.4) node [anchor=north west][inner sep=0.75pt]  [font=\footnotesize]  {$x_{1} ,\ \model( x_{1})$};
\draw (340,275.4) node [anchor=north west][inner sep=0.75pt]  [font=\footnotesize]  {$x_{n} ,\ \model( x_{n})$};
\draw (289,285) node [anchor=north west][inner sep=0.75pt]  [font=\footnotesize]  {$\ldots $};
\draw (554,277.4) node [anchor=north west][inner sep=0.75pt]  [font=\footnotesize]  {$y_{1} ,\ \model( y_{1})$};
\draw (682,277.4) node [anchor=north west][inner sep=0.75pt]  [font=\footnotesize]  {$y_{n} ,\ \model( y_{n})$};
\draw (646,285) node [anchor=north west][inner sep=0.75pt]  [font=\footnotesize]  {$\ldots $};
\draw (375,170.4) node [anchor=north west][inner sep=0.75pt]  [font=\footnotesize]  {$x_{1} ,\ x_{2} ,\ \dotsc \ x_{n}$};
\draw (488.86,225.72) node [anchor=north west][inner sep=0.75pt]  [font=\footnotesize] [align=left] {Target LLM \\ ~~~~~~~~~$\model$};
\draw (191,390.4) node [anchor=north west][inner sep=0.75pt]  [font=\footnotesize]  {$\Delta _{0}^{S}$};
\draw (224.57,390.11) node [anchor=north west][inner sep=0.75pt]  [font=\footnotesize]  {$\Delta _{1}^{S}$};
\draw (382.57,390.11) node [anchor=north west][inner sep=0.75pt]  [font=\footnotesize]  {$\Delta _{\ell}^{S}$};
\draw (318.24,389.92) node [anchor=north west][inner sep=0.75pt]  [font=\footnotesize]  {$\Delta _{i}^{S}$};
\draw (543,393.4) node [anchor=north west][inner sep=0.75pt]  [font=\footnotesize]  {$\Delta _{0}^{T}$};
\draw (576.57,393.11) node [anchor=north west][inner sep=0.75pt]  [font=\footnotesize]  {$\Delta _{1}^{T}$};
\draw (734.57,393.11) node [anchor=north west][inner sep=0.75pt]  [font=\footnotesize]  {$\Delta _{\ell}^{T}$};
\draw (328,467) node [anchor=north west][inner sep=0.75pt]   [align=left] {$\dkw$};
\draw (521,438) node [anchor=north west][inner sep=0.75pt]   [align=left] {$\inbucket$};
\draw (255,184) node [anchor=north west][inner sep=0.75pt]  [font=\footnotesize] [align=left] {\begin{minipage}[lt]{47.18pt}\setlength\topsep{0pt}
\begin{center}
Sample Set \\$\displaystyle S$
\end{center}

\end{minipage}};
\draw (669.57,394.11) node [anchor=north west][inner sep=0.75pt]  [font=\footnotesize]  {$\Delta _{i}^{T}$};
\draw (290,515.67) node [anchor=north west][inner sep=0.75pt]   [align=left] {Output $\accept$ if \( S \) is sampled from \( \model \) else output $\reject$};
\draw (667,206.4) node [anchor=north west][inner sep=0.75pt]  [font=\footnotesize]  {$y_{1} ,\ y_{2} ,\ \dotsc \ y_{n}$};
\draw (262.5,297.61) node [anchor=north west][inner sep=0.75pt]  [color={rgb, 255:red, 173; green, 217; blue, 244 }  ,opacity=1 ] [align=left] {\textcolor[rgb]{0.06,0.08,0.1}{$\bucketing$}};
\draw (612.5,298.61) node [anchor=north west][inner sep=0.75pt]  [color={rgb, 255:red, 173; green, 217; blue, 244 }  ,opacity=1 ] [align=left] {\textcolor[rgb]{0.06,0.08,0.1}{$\bucketing$}};

\end{tikzpicture}
    \caption{Overview of our framework $\alg$. Given a sample set $S$ and a target model $\model$, $\alg$ computes the probability of $S$ under $\model$ and partitions the samples into buckets. $\alg$ then measures the distance between the bucket distribution and the occurrence disparity within each bucket to make a final decision.}
    \label{fig:anubis}\vspace{-1.5em}
\end{figure*}
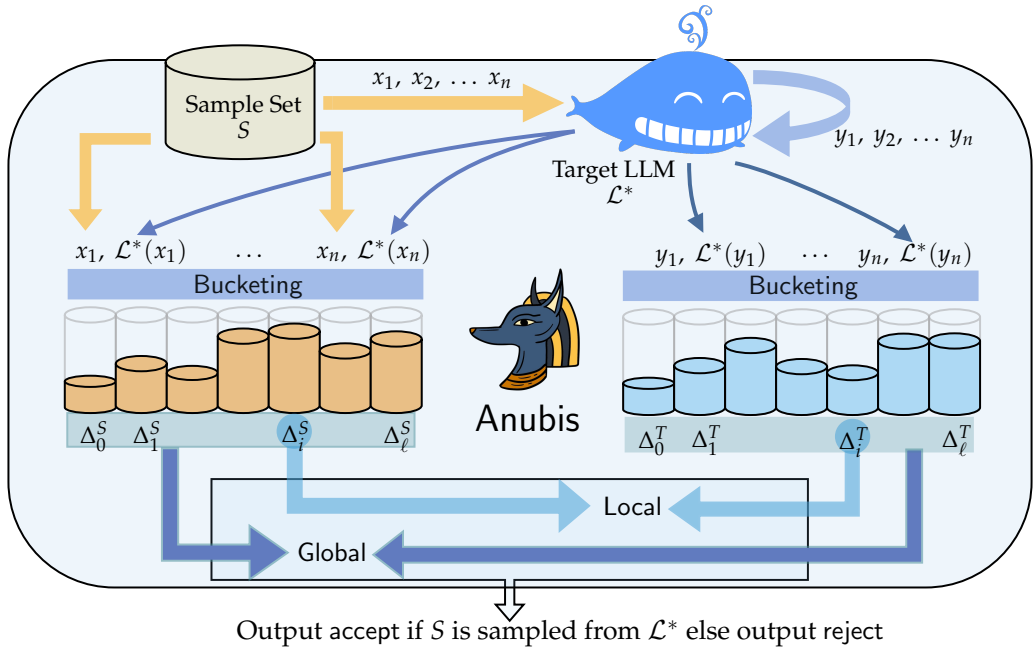

Our problem draws a parallel to the well-established field of authorship attribution~\citep{uchendu2020authorship, uchendu2024topformer}. Traditionally, authorship attribution involves determining the origin of a single sample, often among multiple candidate authors. In our setting, we consider the problem in a multi-sample setting, where evidence can be aggregated from multiple samples. Furthermore, we seek to solve a robust version of the attribution problem, where the dataset may contain samples from a different author LLM. We propose the following problem formalization:

\begin{problem}[Attribution]
Given a set of $n$ samples, a model $\model$, and two thresholds $\lo \%$ and $\hi \%$, the goal is to determine, with high confidence, between two cases:  
\begin{enumerate}
\item Most of the samples ($\geq \hi \%$) are from $\model$.  
\item Very few of the samples ($\leq \lo \%$) are from $\model$.  
\end{enumerate}

\end{problem}

In this paper, we propose $\alg$, a tool for zero-shot attribution that leverages important techniques from distribution testing to address the above. We apply $\alg$ to determine whether most samples in a set originate from a target LLM, selecting \deepseek{}~\citep{GZY+24} or \gemma{}~\citep{team2024codegemma} as the target models. 
To assess the robustness of $\alg$, we construct test datasets using a mix of samples from the target models and \stability{}\citep{PAP+24}.
Our experiments demonstrate that $\alg$ outperforms the state-of-the-art approach $\base$~\citep{mitchell2023detectgpt} in attributing code samples from various LLMs, achieving an AUROC score consistently above 0.9 when $|\hi - \lo| \geq 70$, and delivering surprisingly strong results even with as few as 1000 samples.

\paragraph{Related work} 
There have been a number of works that have studied the problem of identifying LLM-generated content~\citep{wu2025survey}. These works can be broadly divided into three categories: (1) watermarking technology: this approach involves embedding a watermark in the text, which is later extracted by a watermark detector to determine the origin~\citep{gu2022watermarking, kirchenbauer2023reliability, tang2023did}, (2) neural-network based detectors: in this approach a classifier is trained for the task of attribution ~\citep{uchendu2024topformer, kumarage2023j}, and finally (3) zero-shot statistical approach~\citep{mitchell2023detectgpt, bao2023fast}. Our work falls into the third category, where we use distribution testing techniques for the attribution task. 

\paragraph{Organization} We introduce the notation used in this paper and the background in Section~\ref{sec:prelims}. We provide the theoretical basis for our algorithm in~\cref{sec:idealtest}, under some idealized assumptions, before describing how to translate it into a practical algorithm for the task at hand in~\cref{sec:deviations}. Finally, we present the results of our experiments in Section~\ref{sec:experiments}. Complete proofs, additional experiments, and further discussions are available in the Appendix (\ref{sec:appendix}).
\section{Preliminiaries}\label{sec:prelims}

In this work, we are interested in discrete probability distributions over a finite domain $\domain$, which we identify using their probability mass function (pmf), a function $\cP: \Omega \rightarrow [0,1]$ such that $\sum_{x\in \Omega} \cP(x) = 1$. For a subset $S\subseteq\domain$, we let $\cP(S) = \sum_{x\in S}\cP(x)$. Throughout the paper, we will use calligraphic letters to denote distributions, e.g., $\cP, \cL$. 

We define notions of distance between probability distributions as follows: The $\ell_1$ distance between two distributions $\cP$ and $\cP^*$ over $\domain$ is defined as $\|\cP - \cP^*\|_1 = \sum_{x \in \domain} |\cP(x) - \cP^*(x)|$; and the $\ell_\infty$ distance is defined as $\|\cP- \cP^*\|_\infty = \max_{x \in \Sigma^n} |\cP(x) - \cP^*(x)|$. 
Given a set $S \subseteq \domain$ with $\cP(S) > 0$, the \textit{conditional distribution} of $\cP$ on $S$, denoted by $\cond{\cP}{S}$, is a distribution such that for any $x \in S$, $\cond{\cP}{S}(x) = \cP(x)/\cP(S)$. 

\paragraph{Oracles for distributions} We will use \textit{oracles} to formally model interaction with distributions. Hence, to draw independent samples from a distribution $\cP$, we call the $\samp$ oracle. The $\eval$ (for \emph{evaluation}) oracle provides a more powerful type of access, allowing exact evaluation of the probability of a domain element in a distribution, i.e., $\eval(\cP, x) = \cP(x)$. 

In many cases, the exact probabilities are not available, either due to issues arising from finite precision representation or due to the computational complexity of computing exact probabilities. Hence, it is often more realistic to allow for \emph{approximate} evaluation oracles, which return multiplicative approximations of the probability $\cP(x)$. 

\begin{definition}[Approximate \eval oracle~\citep{BatuDKR05,CR:14,BGMV:20}]
    Let $\cP$ be a distribution over $\Omega$, $x \in \Omega$, and $\eta \in [0,1]$. An approximate \eval oracle takes as input $x$ and returns an $\eta$-estimate of $\cP(x)$,  i.e., $\eval(\cP,x) \in [(1-\eta)\cP(x),(1+\eta)\cP(x)]$. The oracle is deterministic and will return the same value for a fixed input $\cP$ and $\eta$.
\end{definition}

\paragraph{Distribution testing} Distribution testing is a well-studied area of study, concerned with questions of the following type: given distribution $\cP$, and a property\footnote{A property is a non-empty collection of distributions.} $\Pi$, the goal is to decide whether $\cP$ satisfies $\Pi$, or is \emph{far} from satisfying $\Pi$. Formally, for a distance parameter $\varepsilon_2$, a confidence parameter $\delta$, with probability at least $1-\delta$, a testing algorithm returns (1) $\accept$ if  $\cP \in \Pi$, or else (2) $\reject$ if $\cP$ is at distance at least $\eps_2$ from $\Pi$ (``is $\eps_2$-far''). We will consider the $\ell_1$ distance in this paper, and the property of interest is $\varepsilon_1$-closeness to a given distribution $\cP^*$. Hence $\Pi = \{\cP \mid  \|\cP - \cP^*\|_1 \leq \varepsilon_1 \}$.
Hence, for input distributions $\cP,\cP^*$, and parameters $\varepsilon_1, \varepsilon_2$ and $\delta$, the test returns (1) $\accept$ if  $\|\cP - \cP^*\|_1 \leq \varepsilon_1$, or else (2) $\reject$ if $\|\cP - \cP^*\|_1 > \varepsilon_2$.

 We will rely on the celebrated Dvoretzky–Kiefer–Wolfowitz (DKW) inequality, which provides a bound on the $\ell_\infty$ distance of an empirical cumulative distribution of samples from the distribution that the samples are drawn from. The optimal constant in the DKW inequality was found by \cite{Massart:90}, and we use it in our analysis.

\begin{lemma}[\cite{DKW:56},\cite{Massart:90}] Let $\widehat{\cP}$ be the empirical distribution generated by $m$ i.i.d.\ samples from a distribution $\cP$ over domain $[n]$. We have that:
$
    \Pr \left[ \max_{i \in [n]}\left|\sum_{j \in [i]}\cP(j)- \sum_{j \in [i]} \widehat{\cP}(j) \right| \geq \varepsilon\right] 
    \leq 2e^{-2m\varepsilon^2} 
$.
\label{lem:dkw}
\end{lemma}

\paragraph{Language Models as Distributions} We start by defining terms related to the domain of language models.
A \textit{token} is an integer from the set $\tokdict = \{1, \ldots, |\Sigma_{tok}|\}$, and a sequence of $n$ tokens is represented as $\sigma_1\sigma_2\ldots \sigma_n$ or $\sigma$ when $n$ is clear from context.
The $k$-length prefix of $\sigma$, denoted as $\sigma_{<k+1}$, refers to the sequence $\sigma_1\sigma_2\ldots \sigma_k$. We will use $\sigma_{<1}$ for the empty sequence. 
A \textit{word} is a symbol from a set of symbols called a \textit{dictionary}, denoted by $\worddict$.  Language models are distributions over sequences of words called \textit{text}.  Formally, a language model $\cL: \worddict^* \to [0,1]$ is a distribution that assigns a probability $\cL(s)$ to every $s \in \worddict^*$. We define \textit{token distribution} $\cLtok: \tokdict^* \to [0,1]$ that assigns a probability $\cLtok(\sigma)$ to every token sequence $\sigma \in \tokdict^*$.

For state-of-the-art LLMs,  next-token prediction is the standard sampling strategy. In the next token prediction, sequence $\sigma$ is sampled from the distribution such that the probability of the $k^{th}$ token in $\sigma$ depends only on the preceding $k$-1 tokens, denoted as $\sigma_{<k}$.
This allows for left-to-right generation where the model samples tokens starting from $\sigma_1$ to $\sigma_n$, predicting one token at a time conditional on the prefix. $\cLtok(\cdot|\sigma_{<i})$ denotes the marginal distribution of the $i^{th}$ token conditioned on the prefix $\sigma_{<i}$. The chain rule of probability lets us factorize $\cLtok(\sigma)$ (the probability of a $n$-token sequence $\sigma$)  as $\cLtok(\sigma) = \prod_{i = 1}^n \cLtok(\sigma_i|\sigma_{<i})$. $\tokdict$ contains a special token $\eos$ to indicate the end-of-string, which is either the last token or is followed by another $\eos$. Formally, $\forall_{i \in [n-1]} \cLtok(\sigma_{i+1} = \eos|\sigma_i = \eos) = 1$.

A tokenizer, consisting of two functions, $\enc$, and $\dec$, serves as an interface between the token space and the word space. We adopt the formal definition of a tokenizer as presented in \cite{gastaldi2024foundations}.

\begin{definition}[Tokenizer]
A tokenizer from dictionary $\worddict$ to token-set $\tokdict$ is defined as a tuple of deterministic mappings, $\tok = (\enc, \dec)$, known as the \textit{encoder} and \textit{decoder}, respectively. The encoder, $\enc: \worddict \to \tokdict^*$, converts a word into a token sequence, while the decoder, $\dec: \tokdict \to \worddict$, converts one token to one word. We can extend the definition of encoder and decoder to sequences of words and tokens, respectively, as $\enc^*: \worddict^* \to \tokdict^*$ and $\dec^*: \tokdict^* \to \worddict^*$.
A tokenizer is called \textit{bijective} if $\enc^*$ is a one-to-one mapping, and $\dec^* = (\enc^*)^{-1}$ over $\enc^*(\worddict^*)$. 
\end{definition}

The process of generating text from an LLM is primarily based on drawing samples from the token distribution $\cLtok$. The tokenizer, specifically the decoder $\dec$, transforms the sampled token sequence into the output text, eventually defining the distribution $\cL$ of the LLM. 
We use Greek lowercase (e.g., $\sigma$) for token sequences and lowercase letters (e.g., $s$) for texts.
\section{An idealised attribution test}\label{sec:idealtest}

In this section, building on (theoretical) results from the distribution testing literature, we present and analyze an algorithm \ to decide whether a distribution $\undertest$ is close to $\ideal$(both defined over the finite domain $\domain$) where we have access to a set of samples from $\undertest$. Furthermore, we have both \samp and \eval oracle access to the reference distribution \ideal. 

\Cref{alg:maintest} takes as input a set of samples $\cS = \{x_1, x_2, \ldots, x_n\}$ from $\undertest$, and oracle access to a distribution $\ideal$, outputs \accept if $\|\undertest - \ideal\|_1 \leq \eps_1$, and \reject if $\|\undertest - \ideal\|_1 \geq \varepsilon_2$. The pseudocode is presented in \cref{alg:maintest}; we outline the main ideas below.

The core idea of \cref{alg:maintest} involves \textit{bucketing}, where a ``bucket'' is a partition of the domain such that all the elements in a bucket have similar probabilities according to $\ideal$. (We will provide the formal definition shortly: for this discussion, it is enough to point out that the number of buckets in the partition is quite small, i.e., logarithmic in the support size, allowing for a sample-efficient tester.) 
The algorithm first draws a fresh set of samples $\cT$ from $\ideal$ and then counts the number of samples of $\cS$ and $\cT$ falling into each bucket. This is done using the $\bucketing$ algorithm (\cref{alg:bucket}).

Intuitively, if there is a big difference in the number of samples from $\cS$ falling into some bucket, as compared to the samples from $\cT$, then this is evidence that $\undertest$ and $\ideal$ differ. To make this rigorous, the algorithm then uses the $\dkw$ test (\cref{alg:DKW}) to determine whether the distribution of samples \emph{over buckets} induced by $\undertest$ and $\ideal$ are consistent.

At this stage, it could still be the case that the distributions induced over buckets are similar for both $\undertest$ and $\ideal$ even though the two distributions significantly differ, as all the discrepancy between the two distributions is hidden ``within the buckets''. To detect this, the
$\inbucket$ algorithm (\cref{alg:inbucket}) is then invoked to test the closeness of the (conditional) distributions within each bucket, using a variant of a chi-squared test for each.

To summarize, the $\dkw$ algorithm tests the closeness of the empirical cumulative distribution functions (CDFs) of the buckets (``global test''), while the $\inbucket$ algorithm tests the closeness of the samples in the buckets (``local tests''). If both tests pass, the algorithm returns \accept; otherwise, it returns \reject.

\begin{algorithm}
    \caption{}%
    \label{alg:maintest}
\begin{algorithmic}[1]    
    \State \textbf{Input}: Set of samples $\cS = \{x_1, x_2, \dots, x_n\}$, and distribution $\ideal$
    \State \textbf{Output}: $\accept$ if $\cS$ is drawn from $\ideal$, else $\reject$ 
    \State  Draw $n$ samples from $\ideal$ and store them in $\cT$
    \State $\Thresh_1 \gets \frac{c_3\eps_2 + \ell\eps_1}{2\ell}$
    \State $\Delta^S \gets \bucketing(\cS, \ideal)$
    \State $\Delta^T \gets \bucketing(\cT, \ideal)$
    \If{$\dkw(\Delta^S, \Delta^T, \Thresh_1)$ returns \reject}
        \State \Return $\reject$
    \EndIf
    \For {$i \in [\ell]$} \Comment{$\Delta^S, \Delta^T$ are lists of size $\ell$}
        \State $\Thresh_2 \gets \left(c_4\eps_2 + 2\ell\eps_1\right)/{2\ell \ideal(\bucket_i)}$
        \If{$\inbucket(\bucket^S_i, \bucket^T_i, \Thresh_2, \cS \cup \cT)$ returns \reject}
            \State \Return $\reject$
        \EndIf
    \EndFor
    \State \Return $\accept$
\end{algorithmic}
\end{algorithm}

\paragraph{Bucketing}
We define the buckets in terms of the probabilities of $\ideal$. We fix the number of buckets by $\ell \eqdef \flr{\log_2 \frac{|\domain|}{c_1\eps_2}} + 1$, and define the buckets as follows, for $1\leq j\leq \ell$:
\begin{equation}
    \label{eq:bucket}
    \bucket_j \eqdef \setOfSuchThat{ x \in \domain }{ 2^{-j} < \ideal(x) \leq 2^{-j+1}}
\end{equation}

We keep the other elements in the ``leftover bucket'' $\bucket_0$, i.e., for all $x \in \bucket_0$, $\ideal(x) \leq \frac{c_1\eps_2}{|\domain|}$. 
The $\bucketing$ algorithm takes as input the distribution $\ideal$ and a set of samples $X$. The algorithm distributes the samples in $\cS$ and $\cT$ into buckets. $\bucketing$ uses the $\eval$ oracle to evaluate the probability of the samples of $X$ in $\ideal$. The bucketing is done based on the probabilities of the samples in $X$ as described in \cref{eq:bucket}.

\begin{algorithm}\small
    \caption{$\bucketing(\ideal, X)$}
    \label{alg:bucket}
\begin{algorithmic}[1]    
    \State \textbf{Input}: Distribution $\ideal$, and sample set $X = \{x_1, x_2, \dots, x_n\}$
    \State \textbf{Output}: List of buckets
    \State Initialise buckets $\bucket_0, \bucket_1, \ldots, \bucket_\ell$
    \For{$x \in  X$}
    \State $j \gets \flr{\log_2(\eval(\ideal, x))} + 1$
    \State $\bucket_j \gets \bucket_j \cup \{x\}$
    \EndFor
    \State \Return $(\bucket_0, \bucket_1, \ldots, \bucket_\ell)$
\end{algorithmic}
\end{algorithm}

\paragraph{The global test (DKW)}
The $\dkw$ algorithm takes as input two list of buckets $(\bucket^S_0, \bucket^S_1, \ldots, \bucket^S_\ell)$ and $(\bucket^T_0, \bucket^T_1, \ldots, \bucket^T_\ell)$, and a threshold $\Thresh$. The algorithm computes the empirical CDF for the two sets of buckets and determines the maximum pointwise difference between the empirical CDFs. If the difference is greater than the threshold $\Thresh$, the algorithm returns \reject. Otherwise it returns \accept (\cref{alg:DKW}).

\paragraph{The Local Test} The $\inbucket$ algorithm takes as input two buckets $\bucket^S_i$ and $\bucket^T_i$, as well as threshold $\Thresh$. The algorithm computes a chi-square-type statistic for each sample in $\bucket^S_i$ and compares their sum, $Z$, to the threshold $\Thresh$. If the sum of $Z$ is less than the threshold $\Thresh$, the algorithm returns \accept. Otherwise it returns \reject (\cref{alg:inbucket}).

\begin{algorithm}[t]
    \caption{$\dkw(\Delta^S, \Delta^T, \Thresh)$}
    \label{alg:DKW}
    \begin{algorithmic}[1]
    \State \textbf{Input}:
    Set of buckets $\Delta^S = (\bucket^S_0, \ldots, \bucket^S_{\ell}), \Delta^T = (\bucket^T_0, \ldots, \bucket^T_\ell)$, threshold $\Thresh$
    \State \textbf{Output}:  \accept or \reject 
    
    \State Set $m \gets |\bigcup_{i=1}^\ell \bucket^S_i|$
    \State Compute the empirical CDF for the two sets of buckets:
    $
       \hat{\Delta}^S(i) = \frac{1}{m}\sum_{j=1}^{i} |\bucket^S_j|,
    \ \ 
        \hat{\Delta}^T(i) = \frac{1}{m}\sum_{j=1}^{i} |\bucket^T_j| 
    $
    \State \textbf{if} \ $  \max_i |\hat{\Delta}^S(i) - \hat{\Delta}^T(i)| > \Thresh$ \ \textbf{then} \ \Return \reject, \ \textbf{else} \ \Return \accept

    \end{algorithmic}
\end{algorithm}

\begin{algorithm}[t]
    \caption{$\inbucket(\bucket^S_i, \bucket^T_i, \Thresh, X)$}
    \label{alg:inbucket}
    \begin{algorithmic}[1]
    \State \textbf{Input}: 
    Buckets $\bucket^S_i$, $\bucket^T_i$, Threshold $\Thresh$, Sample set $X$
    \State \textbf{Output}: \accept or \reject 
    \State $Z \gets 0$
    \For{$x \in X$} 
        \State $z_1 \gets \cnt(x, \bucket^S_i)$, \  $z_2 \gets \cnt(x, \bucket^T_i)$  \Comment{Computing multiplicity of $x$ in $\bucket^S_i, \bucket^T_i$}
        \State $Z \gets Z + \frac{(z_1 - z_2)^2 - z_1 - z_2}{\max(z_1+z_2, 1)}$
    \EndFor
    \State \textbf{if} \ $Z \geq \Thresh$ \ \textbf{then} \ \Return \reject, \ \textbf{else} \ \Return \accept
    \end{algorithmic}
\end{algorithm}

\begin{theorem}[Adapted from~{\cite{CanonneJKL22}}]
    \label{thm:main}
    Let $\ideal$ be a fully specified distribution such that $\|\ideal\|_2 \leq \sqrt{2/|\domain|}$ and $\undertest$ be an unknown distribution, both over the same domain $\domain$. Then there exists an positive absolute constant $c $ such that, for any $0 \leq \eps_2 \leq 1$ and $0 \leq \eps_1 \leq c \eps_2/\ell$, there is an Algorithm that uses 
    $
    \tOh(|\domain|\left({\eps_1}/{\eps_2^2}\right)^2 + |\domain|\left({\eps_1}/{\eps_2^2}\right) + {\sqrt{|\domain|}}/{\eps_2^2})
    $
    many samples to distinguish between $\|\undertest - \ideal\|_1 \leq \eps_1$ and $\|\undertest - \ideal\|_1 \geq \eps_2$ with probability at least 4/5.
\end{theorem}
%
\section{$\alg$: a practical attribution tester}\label{sec:deviations}

To address the problem of attribution within the framework of distribution testing, we adapt \cref{alg:maintest} to construct our attribution tester $\alg$. $\alg$ evaluates whether or not a substantial number of code samples in a set $S$ was generated by a target language model, \model. Given \model and $S$, with high confidence, $\alg$ outputs $\accept$ if most samples are generated by \model, and $\reject$ otherwise. In distribution testing terms, this reduces to determining whether $S$ comes from a distribution close to or distant from \model.

The primary challenges of designing $\alg$ are (1) the infeasibility of directly accessing the probabilities $\model(s)$, which are required for the \eval oracle. We circumvent this by designing the \evalplus algorithm that estimates the probability of a sample $s$ generated by a language model \model.
(2) Determining the number of buckets $\ell$ is a significant bottleneck because, ideally, this requires knowledge about the domain size of $\model$. We circumvent this with the Monte Carlo approach, wherein we draw samples to estimate the number of buckets while incurring some error, which is accounted for in the analysis. To address this, we introduce a parameter called the \textit{leftover fraction} $\bw$, which ensures that no more than a $\bw$ fraction of the samples fall into the leftover bucket $\Delta_0$.

\subsection{Designing \eval oracles for language models}\label{sec:evalplus}
The \eval oracle is designed to estimate the probability of a text being generated by a language model. Specifically, given a model $\cL$ and a text $s$, \eval computes the probability $\cL(s)$ that $s$ was produced by $\cL$. However, designing an \eval oracle for language models is challenging due to the non-invertible nature of tokenizers.

\paragraph{Non-invertibility of Tokenizers}  Tokenizers are typically non-injective functions~\citep{sennrich2015neural}, meaning different token sequences, such as $a_1a_2a_3a_4$ and $b_1b_2b_3$ from the token-set $\tokdict$, can map to the same text $s_1s_2$, resulting in a collision. For instance, the tokenizer used by \gpt{}~\citep{radford2019language} maps both token sequences $64\circ 14512\circ 65$ and $64\circ 0 \circ 28 \circ 65$ to the text `a!=b'. 
This non-invertibility complicates the direct computation of $\cL(s)$. While the \eval oracle provides easy access to the marginal probabilities $\cLtok(\sigma_i | \sigma_{<i})$ for a token sequence $\sigma$, calculating $\cL(s)$ from these probabilities remains a non-trivial task unless these collisions are known in advance.

To address this, we propose the algorithm $\evalplus$, which leverages the \eval oracle for the marginal probabilities $\cLtok(\sigma_i | \sigma_{<i})$ to estimate the likelihood that $s$ was generated by the language model $\cL$.
To evaluate $\cL(s)$ we provide $\evalplus$ with the tokenizer $\tok$ used by the model $\cL$, token distribution $\cLtok$, the token sequence $\sigma := \sigma_1\sigma_2\ldots \sigma_n$ such that $\sigma = \enc(s)$, and the depth parameter $d$. If $\sigma$'s length is less than $d$, the algorithm iterates over the tokens, computing the probability of each $\sigma_i$ in $\cLtok(\cdot|\sigma_{<i})$ using the \eval oracle. However, as discussed earlier, the tokenization process can lead to multiple token sequences that decode the same text. To account for this, we introduce the function $\getcoll$ that takes a tokenizer $\tok$, a token sequence $\sigma$, and the depth parameter $d$ as input and returns the set of token sequences of length $d$ that decode to the same text as $\dec(\sigma)$. $\evalplus$ then computes the probability of each token sequence in the set returned by $\getcoll$ and adds it to the probability of the token in the token sequence to improve the estimate of the probability $\cL(s)$.
If $\sigma$'s length exceeds $d$, $\evalplus$ splits the sequence into two parts and recursively processes each part.

%
%

\begin{algorithm}[ht]
	\caption{$\evalplus(\cLtok, \tok, \sigma, d)$}
	\label{alg:evalplus}
	\begin{algorithmic}[1]    
		\State \textbf{Input}: Token distribution $\cLtok$, tokenizer $\tok$, token string $\sigma$, depth parameter $d\geq 1$
		\State \textbf{Output}:  A lower bound on the value of $\cL(s)$
		\If{$|\sigma| \leq d$}
		\State $Coll \gets \getcoll(\tok, \sigma, d)$, \ $prob \gets 0$
		\For {$\sigma' \in Coll$}
		\State $prob \gets prob + \eval(\cLtok, \sigma') $
		\EndFor
		\State  \Return $prob$
		\EndIf
		\State \Return $\sum_{i=1}^{d} (\evalplus(\cLtok, \tok, \sigma_1 \sigma_2 \ldots \sigma_{i}, i) 
		\, \cdot \evalplus(\cLtok, \tok, \sigma_{i+1} \sigma_{i+2} \ldots \sigma_n, d))$ 
	\end{algorithmic}
\end{algorithm}

\section{Experiments}\label{sec:experiments}

To assess the practical effectiveness of our algorithm, we implemented $\alg$ to test attribution on a dataset of code samples generated by LLMs. Our goal is to investigate the following research questions:

\textbf{RQ1} For a target model \model, parameters $\lo\%$, $\hi\%$, and a set of code samples $S$, how accurately can $\alg$ determine whether 1) fewer than $\lo\%$, or 2) more than $\hi\%$ of the samples were drawn from \model?

\textbf{RQ2} How does the performance of $\alg$ change with varying number of samples?

\paragraph{Models} As our objective revolves around testing attribution in a set of code samples, we focus on popular code LLMs. 
In our experimental setup, we employed three LLMs, each chosen for a specific role within the setup. The selected versions of the models were of similar size, with parameter counts ranging from 1  to 3 billion. 
(1) We used the 3-billion-parameter version of \stability{}~\citep{PAP+24} to generate the primary code dataset. 
(2) The 1.3-billion-parameter version of \deepseek{}~\citep{GZY+24} and (3) the 2-billion-parameter version of \gemma{}~\citep{team2024codegemma} were employed as the target LLMs. These models were chosen as target models because of their smaller size which allowed us to run experiments more efficiently.

\begin{figure}[t]\vspace{-0.5em}
  \centering
  \includegraphics[width=\linewidth]{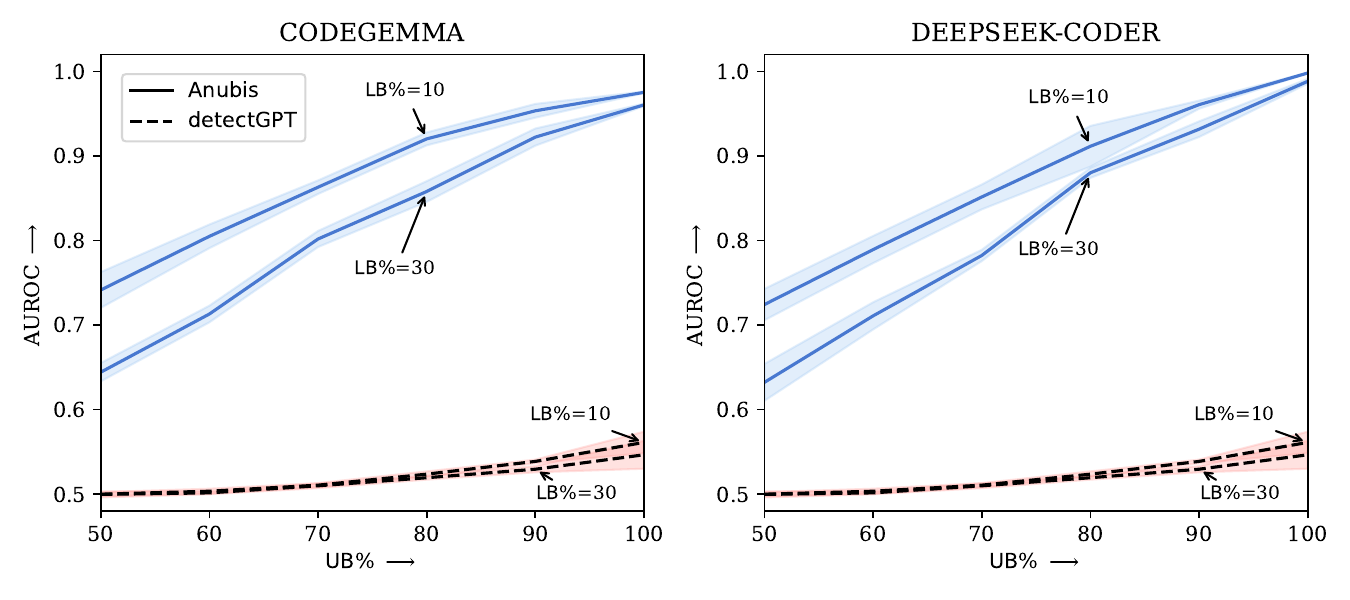}\vspace{-2em}
  \caption{The AUROC of $\alg$ and $\base$ was evaluated against the target LLMs -- \deepseek{} and \gemma{}. The AUROC values are averaged over multiple runs across different datasets with the same contamination level $\gamma\%$. The shaded regions indicate the uncertainty across independent runs. The dashed line represents $\base$, while the solid line represents $\alg$.}
  \label{fig:auroc}
\end{figure}\vspace{-0.5em}

\paragraph{Datasets} 
The primary requirement for our evaluation is the availability of code samples from different sources. To generate the code datasets, we adopted a generate-and-sanitize approach. Specifically, we used prompts from the \textsc{HumanEval} dataset~\citep{CTJY+21}, which contains 164 Python programming problems, each accompanied by test cases. These problems require knowledge of basic mathematics and can be solved with a few lines of code. For each prompt, we queried three LLMs to generate 2K samples, resulting in three datasets, each comprising 164 tasks with 2K samples per task. Additionally, we constructed an `Adversarial Set` by generating equivalent datasets using the target LLMs. 
To evaluate $\alg$'s robustness in attribution, we simulated contamination by replacing a fraction of the original samples with adversarial ones. For each contamination level $\gamma\%$ ($\gamma \in \{0, 10, \ldots, 90\}$) and each fresh dataset, we randomly replaced $\gamma\%$ of the samples with samples from the adversarial set. We used these contaminated datasets to assess $\alg$'s performance in detecting the presence of samples from the target LLMs.

\paragraph{Comparison}
We compared $\alg$ against the SOTA zero-shot statistical approach for identifying LLM-generated text: $\base$~\citep{mitchell2023detectgpt}. $\base$ operates on the principle that LLM-generated samples tend to be drawn from the local maxima of the target LLM's distribution, whereas text generated from other sources may originate from a flatter region of the same target LLM's distribution. 
Since $\base$ is designed for use on individual code samples, we extended it for dataset-level analysis. $\base$ computes a score called {perturbation discrepancy} for each sample, using which it takes a binary decision for that sample. To make $\base$ work for the dataset, we aggregated the scores for all samples to evaluate the overall attribution for the dataset.

\paragraph{Evaluation metric} Following the setup in~\cite{mitchell2023detectgpt}, we used the Area Under the Receiver Operating Characteristic Curve (AUROC) as our evaluation metric. AUROC measures the ability of the algorithm to distinguish between two classes, in this case, samples generated by the target LLMs and samples that are not. An AUROC of 1 indicates perfect performance, while 0.5 is equivalent to random guessing. In our experiments, we applied $\alg$ to determine whether the proportion of samples generated by the target LLMs within a set $S$ was below $\lo\%$ or above $\hi\%$. Accordingly, our positive class was defined as sets containing more than $\hi\%$ of samples from the target LLMs, while the negative class was sets containing fewer than $\lo\%$ of samples from the target LLMs.

\paragraph{Hyperparameters}\label{sec:params} 
The key hyperparameters of $\alg$ are: (1) the leftover fraction $\bw$, which is fixed at 0.05 in all experiments related to \cref{fig:auroc,fig:nsamp}. In the appendix, we present an additional experiment to demonstrate the low sensitivity of $\alg$ to $\bw$.  
(2) To maintain the algorithm's tractability, we set the depth parameter $d$ in $\evalplus$ to 2.

%
%
%

%
%
%
%
%
%
%
%
%

%
%
%
%
%
%

%
%
%
%
%
%

\subsection{Experiment results and discussion}

  \paragraph{\textbf{RQ1}}
  For each of the 164 prompts in \textsc{HumanEval}, we applied $\alg$ and $\base$ to determine whether a dataset corresponding to the prompt contained fewer than $\lo\%$ or more than $\hi\%$ of samples from the target LLMs (\deepseek{} and \gemma{}).
  The results, summarized in \cref{fig:auroc}, show the $\hi\%$ on the x-axis, ranging from 50 to 100 in increments of 10, and the AUROC on the y-axis. The result clearly shows that $\alg$ significantly outperforms $\base$ in all cases. The AUROC values rise as the gap between \lo and \hi widens, surpassing 0.9 for $\hi = \{90,100\}$, against $\lo = \{10, 20, 30\}$, across all the models we tested, e.g., \{\deepseek{}, \gemma{}\}. This confirms that the task becomes easier for both testers ($\alg$ and $\base$) as the tolerance gap, that is $|\hi-\lo|$, widens.
  We anticipated $\base$ to perform poorly in this setting for the following key reason: the samples under evaluation are code samples, resulting in a sparse underlying distribution and are likely drawn from local maxima of the target LLM distribution due to syntactic and semantic constraints. Since $\base$'s working principle primarily revolves around the assumption that samples sourced from other models belong to flatter (not a local maxima) regions of the target distribution, it is expected to fail in this scenario.

\begin{figure}[t]
  \centering
  \includegraphics[width=\linewidth]{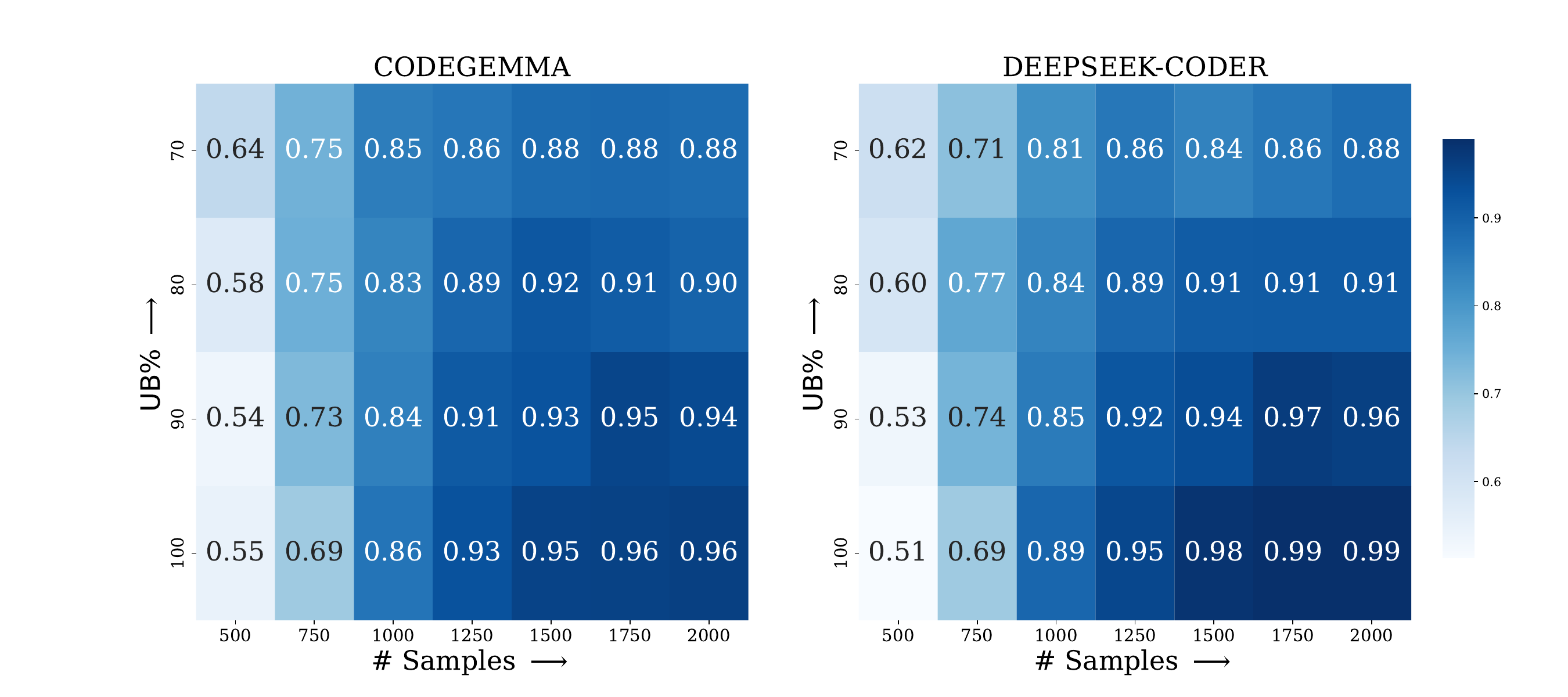}
  \vspace{-2em}
  \caption{The AUROC of $\alg$ when the number of samples is varied from 500 to 2000, evaluated against \deepseek{} and \gemma{}. $\lo\%$ is set to 0 in all cases. The colour intensity represents the AUROC value, with darker shades indicating higher AUROC.}
  \label{fig:nsamp}
\end{figure}

  \paragraph{\textbf{RQ2}} 
  We also studied the effect of the number of samples in the dataset on the performance of $\alg$. We varied the number of samples within \{500, 750, 1000, 1250, 1500, 1750, 2000\} and evaluated the performance of $\alg$.  
  The results, presented in \cref{fig:nsamp}, show that increasing the number of samples improves the decision-making quality of $\alg$, which is natural in sampling-based algorithms. Notably, despite the theoretical analysis of $\alg$ suggesting the need for a large number of samples, potentially comparable to the vocabulary size of the target LLM, the practical results indicate that $\alg$ achieves at least 80\% reliability with as few as 1000 samples. This highlights the robustness of $\alg$ even with small sample sizes, reflecting the strong theoretical foundation.

\paragraph{Limitations}
$\alg$ assumes access to likelihood of samples, and this limits the use of the model to scenarios where the log-probabilities are available for query. Likelihood is available with certain APIs, and if the model is run locally.  
\section*{Acknowledgements}
Clément L. Canonne Supported by an ARC DECRA (DE230101329). Uddalok Sarkar is supported by the Google PhD Fellowship. This research is part of the programme DesCartes and is supported by the National Research Foundation, Prime Minister’s Office, Singapore, under its Campus for Research Excellence and Technological Enterprise (CREATE) programme. The computational works of this article were performed on the resources of the National Supercomputing Centre, Singapore \url{https : //www.nscc.sg}.

\appendix

\newpage

\section{Appendix}\label{sec:appendix}
\subsection{Details of the proof of~\cref{thm:main}}
We first restate the theorem:
\begin{theorem}[Adapted from~{\cite{CanonneJKL22}}]
    \label{thm:main:restated}
    Let $\ideal$ be a fully specified distribution such that $\|\ideal\|_2 \leq \sqrt{2/|\domain|}$ and $\undertest$ be an unknown distribution, both over the same domain $\domain$. Then there exists an positive absolute constant $c $ such that, for any $0 \leq \eps_2 \leq 1$ and $0 \leq \eps_1 \leq c \eps_2/\ell$, there is an Algorithm that uses $$\tOh\left(|\domain|\left(\frac{\eps_1}{\eps_2^2}\right)^2 + |\domain|\left(\frac{\eps_1}{\eps_2^2}\right) + \frac{\sqrt{|\domain|}}{\eps_2^2}\right)$$ many samples to distinguish between $\|\undertest - \ideal\|_1 \leq \eps_1$ and $\|\undertest - \ideal\|_1 \geq \eps_2$ with probability at least 4/5. (Moreover, a detailed bound can be found in \cref{eq:sample:n1:restated,eq:sample:n2:restated}.) 
\end{theorem}
The beginning of the proof is given in the submission; we reproduce it here for completeness, but the changes start at~\cref{eq:sample:n2:restated}.
\begin{proof}
Let us assume that the buckets $\bucket_0, \ldots, \bucket_\ell$ are defined as in \cref{eq:bucket}. By~\cite[Lemma~C.7]{CanonneJKL22}, if $\|{\undertest} - {\ideal}\|_1 \leq \eps_1$, then
\begin{enumerate}
    \item $\undertest(\bucket_0) - \ideal(\bucket_0) \leq \eps_1$ 
    \item $|\undertest(\bucket_j) - \ideal(\bucket_j)| \leq \eps_1$ for all $1\leq j\leq \ell$
    \item $\|\cond{\undertest}{\bucket_j}- \cond{\ideal}{\bucket_j}\|_1 \leq \frac{2\eps_1}{\ideal(\bucket_j)}$ for all $1\leq j\leq \ell$
\end{enumerate}

\noindent By~\cite[Lemma~C.8]{CanonneJKL22}, conversely, for any choice of constants $c_1,c_2,c_3,c_4 > 0$, if
\begin{enumerate}
    \item $\undertest(\bucket_0) \leq c_2\eps_2$ 
    \item $|\undertest(\bucket_j) - \ideal(\bucket_j)| \leq \frac{c_3\eps_2}{\ell}$ for all $1\leq j\leq \ell$
    \item $\|\cond{\undertest}{\bucket_j} - \cond{\ideal}{\bucket_j}\|_1 \leq \frac{c_4\eps_2}{\ell\cdot \ideal(\bucket_j)}$
\end{enumerate}
then $\|{\undertest} - {\ideal}\|_1 \leq (c_1 + c_2 + 2c_3 + c_4)\eps_2$.\medskip

This gives the following constraints on $c_1,c_2$ and (looking ahead) probabilities of failure $\delta_1, \delta_2$:
\begin{align}
    c_1 + c_2 + \frac{1}{2}c_3 + c_4 &\leq c \label{eq:c1:restated}\\
    \eps_1 &\leq (c_2 - c_1)\cdot \eps_2\\
    \eps_1 &\leq c_3\cdot \frac{\eps_2}{\ell} \label{eq:11}\\
    2\eps_1 &\leq c_4\cdot \frac{\eps_2}{\ell} \label{eq:12}\\
    \delta_1 + \ell \cdot \delta_2 &\leq \frac{1}{5} 
\end{align}
We will also add the condition 
\begin{align}
c_4 \geq 2c_3
\end{align}
on the constants, which will be used in the analysis (which uses $\frac{c_4\eps_2}{2\ell} \geq \frac{2c_3\eps_2}{\ell}$). Note that this condition, combined with~\eqref{eq:12}, implies~\eqref{eq:11}.
\begin{enumerate}
    \item For $\bucket_0$ we want to distinguish between $|\undertest(\bucket_0) - \ideal(\bucket_0)| \leq \eps_1$ from $|\undertest(\bucket_0) - \ideal(\bucket_0)| \geq (c_2 - c_1)\eps_2$. And for every subsequent buckets $1\leq j\leq \ell$, we need to distinguish $|\undertest(\bucket_j) - \ideal(\bucket_j)| \leq \eps_1$ from $|\undertest(\bucket_j)-\ideal(\bucket_j)| > \frac{c_3\eps_2}{\ell}$ with error probability $\delta_1$ (overall). This can be done by empirically estimating the distributions $\undertest(\bucket_j)$ and $\ideal(\bucket_j)$ (where $j = 0, 1, \ldots \ell$) within the bound $\pm \eta \eqdef \min\left(\frac{(c_2-c_1)\eps_2 - \eps_1}{4},\frac{c_3\eps_2-\ell\eps_1}{4\ell}\right)$.
    From \cref{lem:dkw}, this would require 
    \begin{equation}
        \label{eq:sample:n1:restated}
        n_1 = \clg{ \frac{8}{\eta^2} \ln \frac{4}{\delta_1} }
    \end{equation}
    samples from each of the distributions $\undertest$ and $\ideal$.
    
    \item For every $1\leq j\leq \ell$, we need to distinguish $\|\cond{\undertest}{\bucket_j}-\cond{\ideal}{\bucket_j}\|_1 \leq \frac{2\eps_1}{\ideal(\bucket_j)}$ from $\|\cond{\undertest}{\bucket_j}-\cond{\ideal}{\bucket_j}\|_1 > \frac{c_4\eps_2}{\ell\cdot \ideal(\bucket_j)}$ with error probability $\delta_2$ (each). To do so, we take 
    \begin{equation}
        \label{eq:sample:n2:restated}
            n_2 = \left\lceil C\cdot \left(|\domain|\max\!\left(\left(\tfrac{\eps_1}{\eps_2^2}\right)^2,\left(\tfrac{\eps_1}{\eps_2^2}\right)\right) + \tfrac{\sqrt{|\domain|}}{\eps_2^2}\right) \ln \ell  \right\rceil
    \end{equation}
    samples in total, where $C=C(c_1,c_2,c_3,c_4)$ is an absolute constant, and call the $\ell$ identity testing algorithms on each of the distributions, with the above parameter (for $\eps_2$) using for the $j$-th the samples which fell in $\bucket_j$. The details of the above bound follow the same outline as~\cite{CanonneJKL22}: we provide them here for completeness. 

    For any $1\leq j\leq \ell$ such that $\ideal(\bucket_j)\geq \frac{c_4\eps_2}{2\ell} \geq \frac{2c_3\eps_2}{\ell}$,\footnote{We only need to consider the $j$s such that $\ideal(\bucket_j)\geq \frac{c_4\eps_2}{2\ell}$, as for the others the condition $\|\cond{\undertest}{\bucket_j}-\cond{\ideal}{\bucket_j}\|_1 > \frac{c_4\eps_2}{\ell\cdot \ideal(\bucket_j)}$ can never be satisfied (the RHS being greater than 2) and so the test is trivial.} then we can assume that $\undertest(\bucket_j)\geq \frac{1}{2}\ideal(\bucket_j)$, otherwise
    \[
        |\undertest(\bucket_j) - \ideal(\bucket_j)| > \frac{1}{2}\ideal(\bucket_j) \geq \frac{c_3\eps_2}{\ell}
    \]
    and the test from 2) will reject anyway. Now, by a Chernoff bound, taking $n_2$ samples from $\undertest$ we will get at least
    \[
        n_{2,j} \coloneqq n_2\cdot \frac{\undertest(\bucket_j) }{2}
    \]
    samples from $\cond{\undertest}{\bucket_j}$ with probability at least $1-e^{-c_3\eps_2 n_2/\ell} \geq 1-\frac{\delta_2}{2}$, as long as 
    \[
    n_2 \geq \frac{\ell}{c_3\eps_2} \ln \frac{2}{\delta_2}
    \]
    (and we can use the same $n_2$ samples for all at most $\ell$ such intervals, by a union bound). We use these $n_{2,j}$ samples for the \textsf{Local} test on the $j$-th bucket, with parameters
    \[
        \eps_2' \eqdef \frac{c_4\eps_2}{\ell\ideal(\bucket_j)},\quad \eps_1'\eqdef \frac{2\eps_1}{\ideal({\bucket_j})}
    \]
    which is guaranteed to be accurate with probability $\delta_2/2$ as long as
    \[
        n_{2,j} \geq C\cdot \left(|\bucket_j|\left(\frac{{\eps'}_1}{{\eps'}_2^2}\right)^2 + |\bucket_j|\left(\frac{{\eps'}_1}{{\eps'}_2^2}\right) + \frac{\sqrt{|\bucket_j|}}{{\eps'}_2^2}\right) \ln\frac{2}{\delta_2}
    \]
    (from Theorem~2.1 of~\cite{CanonneJKL22:arxiv}). Since 
    $
    \ideal({\bucket_j})  \leq 1
    $
    and $|\bucket_j| \leq |\domain|$, the condition from above, which from the settings of $\eps_1', \eps_2'$ can be rewritten as
    \begin{align*}
        n_{2,j} \geq C\cdot \left(|\bucket_j|\cdot \frac{4\ell^2\ideal({\bucket_j})^2}{c_4^2} \cdot \left(\frac{{\eps}_1}{{\eps}_2^2}\right)^2 + |\bucket_j|\cdot \frac{2\ell\ideal({\bucket_j})}{c_4} \cdot \left(\frac{{\eps'}_1}{{\eps'}_2^2}\right)\right . \\ 
        \left . + \frac{\sqrt{|\bucket_j|}}{{\eps'}_2^2}\cdot \frac{\ell^2\ideal({\bucket_j})^2}{c_4^2} \cdot \right) \ln\frac{2}{\delta_2}
    \end{align*}
    gives the (stronger, sufficient) condition,
    \[
        n_{2,j} \geq C\cdot \frac{\ideal(\bucket_j)}{4}\cdot  \left(\frac{16\ell^2}{c_4^2} \cdot |\domain| \cdot  \left(\frac{{\eps}_1}{{\eps}_2^2}\right)^2 + \frac{8\ell}{c_4}\cdot |\domain| \cdot \left(\frac{{\eps'}_1}{{\eps'}_2^2}\right) + \frac{4\ell^2}{c_4^2} \cdot \frac{\sqrt{|\domain|}}{{\eps'}_2^2}\right) \ln\frac{2}{\delta_2}\,. \tag{$\ast$}
    \]
    But as 
    \[
    n_{2,j} = n_2\cdot \frac{\undertest(\bucket_j) }{2} \geq n_2\cdot \frac{\ideal(\bucket_j) }{4}
    \]
    fo the buckets considered, the condition $(\ast)$ holds simultaneously for all these buckets with probability at least $1-\delta_2/2$, as long as (1)~$n_2 \geq \frac{\ell}{c_3\eps_2} \ln \frac{2}{\delta_2}$, and (2)~$n_2$ satisfies
    \[
        n_{2} \geq C\cdot \left(\frac{16\ell^2}{c_4^2} \cdot |\domain| \cdot  \left(\frac{{\eps}_1}{{\eps}_2^2}\right)^2 + \frac{8\ell}{c_4}\cdot |\domain| \cdot \left(\frac{{\eps'}_1}{{\eps'}_2^2}\right) + \frac{4\ell^2}{c_4^2} \cdot \frac{\sqrt{|\domain|}}{{\eps'}_2^2}\right) \ln\frac{2}{\delta_2}.
    \]
    That is, for this step to be successful with probability at least $1-\ell\cdot \delta$ overall by a union bound, it suffices to have
    \begin{align}
        n_{2} \geq \max\left(\frac{\ell}{c_3\eps_2} , C\cdot \left(\frac{16\ell^2}{c_4^2} \cdot |\domain| \cdot  \left(\frac{{\eps}_1}{{\eps}_2^2}\right)^2 + \frac{8\ell}{c_4}\cdot |\domain| \cdot \left(\frac{{\eps'}_1}{{\eps'}_2^2}\right) + \frac{4\ell^2}{c_4^2} \cdot \frac{\sqrt{|\domain|}}{{\eps'}_2^2}\right)\right) \ln\frac{2}{\delta_2}
    \end{align}
\end{enumerate}
\end{proof}

\subsection{Handling tokenizer collisions with $\getcoll$}

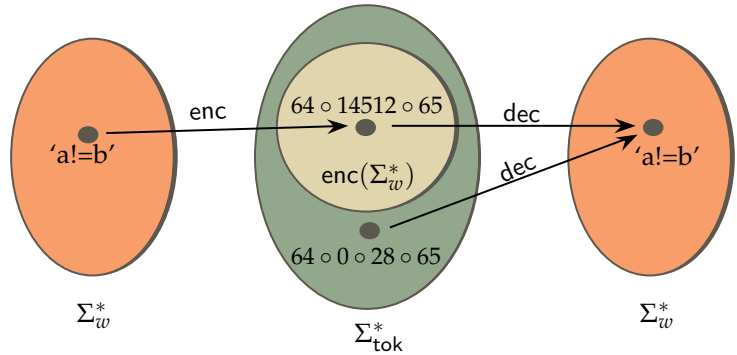
\begin{figure}[b]
    \centering

\tikzset{every picture/.style={line width=0.75pt}} %

\begin{tikzpicture}[x=0.75pt,y=0.55pt,yscale=-1,xscale=1]
\draw  [color={rgb, 255:red, 91; green, 88; blue, 79 }  ,draw opacity=1 ][fill={rgb, 255:red, 247; green, 158; blue, 107 }  ,fill opacity=1 ][general shadow={fill={rgb, 255:red, 91; green, 88; blue, 79 }  ,shadow xshift=1.5pt,shadow yshift=0pt, opacity=1 }] (145.33,133.5) .. controls (145.33,88.49) and (163.69,52) .. (186.33,52) .. controls (208.98,52) and (227.33,88.49) .. (227.33,133.5) .. controls (227.33,178.51) and (208.98,215) .. (186.33,215) .. controls (163.69,215) and (145.33,178.51) .. (145.33,133.5) -- cycle ;
\draw  [color={rgb, 255:red, 91; green, 88; blue, 79 }  ,draw opacity=1 ][fill={rgb, 255:red, 146; green, 167; blue, 140 }  ,fill opacity=1 ][general shadow={fill={rgb, 255:red, 91; green, 88; blue, 79 }  ,shadow xshift=0.75pt,shadow yshift=0pt, opacity=1 }] (268.33,133.5) .. controls (268.33,75.79) and (293.63,29) .. (324.83,29) .. controls (356.04,29) and (381.33,75.79) .. (381.33,133.5) .. controls (381.33,191.21) and (356.04,238) .. (324.83,238) .. controls (293.63,238) and (268.33,191.21) .. (268.33,133.5) -- cycle ;
\draw  [color={rgb, 255:red, 91; green, 88; blue, 79 }  ,draw opacity=1 ][fill={rgb, 255:red, 224; green, 213; blue, 173 }  ,fill opacity=1 ][general shadow={fill={rgb, 255:red, 91; green, 88; blue, 79 }  ,shadow xshift=1.5pt,shadow yshift=0pt, opacity=1 }] (279.33,113) .. controls (279.33,80.97) and (299.48,55) .. (324.33,55) .. controls (349.19,55) and (369.33,80.97) .. (369.33,113) .. controls (369.33,145.03) and (349.19,171) .. (324.33,171) .. controls (299.48,171) and (279.33,145.03) .. (279.33,113) -- cycle ;
\draw  [color={rgb, 255:red, 91; green, 88; blue, 79 }  ,draw opacity=1 ][fill={rgb, 255:red, 247; green, 158; blue, 107 }  ,fill opacity=1 ][general shadow={fill={rgb, 255:red, 91; green, 88; blue, 79 }  ,shadow xshift=1.5pt,shadow yshift=0pt, opacity=1 }] (426.33,133.5) .. controls (426.33,88.49) and (444.69,52) .. (467.33,52) .. controls (489.98,52) and (508.33,88.49) .. (508.33,133.5) .. controls (508.33,178.51) and (489.98,215) .. (467.33,215) .. controls (444.69,215) and (426.33,178.51) .. (426.33,133.5) -- cycle ;
\draw  [color={rgb, 255:red, 91; green, 88; blue, 79 }  ,draw opacity=1 ][fill={rgb, 255:red, 91; green, 88; blue, 79 }  ,fill opacity=1 ][line width=3]  (181,118.17) .. controls (181,116.42) and (182.42,115) .. (184.17,115) .. controls (185.92,115) and (187.33,116.42) .. (187.33,118.17) .. controls (187.33,119.92) and (185.92,121.33) .. (184.17,121.33) .. controls (182.42,121.33) and (181,119.92) .. (181,118.17) -- cycle ;
\draw  [color={rgb, 255:red, 91; green, 88; blue, 79 }  ,draw opacity=1 ][fill={rgb, 255:red, 91; green, 88; blue, 79 }  ,fill opacity=1 ][line width=3]  (321,113.17) .. controls (321,111.42) and (322.42,110) .. (324.17,110) .. controls (325.92,110) and (327.33,111.42) .. (327.33,113.17) .. controls (327.33,114.92) and (325.92,116.33) .. (324.17,116.33) .. controls (322.42,116.33) and (321,114.92) .. (321,113.17) -- cycle ;
\draw  [color={rgb, 255:red, 91; green, 88; blue, 79 }  ,draw opacity=1 ][fill={rgb, 255:red, 91; green, 88; blue, 79 }  ,fill opacity=1 ][line width=3]  (323,184.17) .. controls (323,182.42) and (324.42,181) .. (326.17,181) .. controls (327.92,181) and (329.33,182.42) .. (329.33,184.17) .. controls (329.33,185.92) and (327.92,187.33) .. (326.17,187.33) .. controls (324.42,187.33) and (323,185.92) .. (323,184.17) -- cycle ;
\draw  [color={rgb, 255:red, 91; green, 88; blue, 79 }  ,draw opacity=1 ][fill={rgb, 255:red, 91; green, 88; blue, 79 }  ,fill opacity=1 ][line width=3]  (466,113.17) .. controls (466,111.42) and (467.42,110) .. (469.17,110) .. controls (470.92,110) and (472.33,111.42) .. (472.33,113.17) .. controls (472.33,114.92) and (470.92,116.33) .. (469.17,116.33) .. controls (467.42,116.33) and (466,114.92) .. (466,113.17) -- cycle ;
\draw [color={rgb, 255:red, 0; green, 0; blue, 0 }  ,draw opacity=1 ]   (193.33,117) -- (312.34,112.12) ;
\draw [shift={(315.33,112)}, rotate = 177.65] [fill={rgb, 255:red, 0; green, 0; blue, 0 }  ,fill opacity=1 ][line width=0.08]  [draw opacity=0] (10.72,-5.15) -- (0,0) -- (10.72,5.15) -- (7.12,0) -- cycle    ;
\draw [color={rgb, 255:red, 0; green, 0; blue, 0 }  ,draw opacity=1 ]   (337.33,112) -- (457.33,112) ;
\draw [shift={(460.33,112)}, rotate = 180] [fill={rgb, 255:red, 0; green, 0; blue, 0 }  ,fill opacity=1 ][line width=0.08]  [draw opacity=0] (10.72,-5.15) -- (0,0) -- (10.72,5.15) -- (7.12,0) -- cycle    ;
\draw [color={rgb, 255:red, 0; green, 0; blue, 0 }  ,draw opacity=1 ]   (336.33,182) -- (458.66,119.37) ;
\draw [shift={(461.33,118)}, rotate = 152.89] [fill={rgb, 255:red, 0; green, 0; blue, 0 }  ,fill opacity=1 ][line width=0.08]  [draw opacity=0] (10.72,-5.15) -- (0,0) -- (10.72,5.15) -- (7.12,0) -- cycle    ;

\draw (177,231.4) node [anchor=north west][inner sep=0.75pt]    {$\worddict^*$};
\draw (461,231.4) node [anchor=north west][inner sep=0.75pt]    {$\worddict^*$};
\draw (300.33,135.4) node [anchor=north west][inner sep=0.75pt]    {$\enc( \worddict^* )$};
\draw (317,245.4) node [anchor=north west][inner sep=0.75pt]    {$\tokdict^*$};
\draw (163.17,122.57) node [anchor=north west][inner sep=0.75pt]    {`a!=b'};
\draw (285,89.57) node [anchor=north west][inner sep=0.75pt]    {\small $64\circ 14512\circ 65$};
\draw (458.17,123.57) node [anchor=north west][inner sep=0.75pt]    {`a!=b'};
\draw (285,193.57) node [anchor=north west][inner sep=0.75pt]    {\small $64\circ 0\circ 28 \circ 65$};
\draw (233.58,98.11) node [anchor=north west][inner sep=0.75pt]  [rotate=-356.66]  {$\enc$};
\draw (392.06,94.31) node [anchor=north west][inner sep=0.75pt]  [rotate=-0.42]  {$\dec$};
\draw (387.87,138.58) node [anchor=north west][inner sep=0.75pt]  [rotate=-335.37]  {$\dec$};

\end{tikzpicture}
    \caption{Tokenizers are non-invertible. The GPT2 tokenizer encodes the string ``a!=b'' to a token sequence, however there are two  sequences that decode to the same string, demonstrating a collision.}
    \label{fig:tokenizer}
\end{figure}

We present a detailed visualization illustrating the relationships between $\worddict^*$, $\tokdict^*$ and $\enc(\worddict^*)$ are related. Typically, $\enc(\worddict^*)$ forms a proper subset of $\tokdict^*$, while the domain of the function $\dec$ is $\worddict^*$. Due to the non-invertibility of tokenizers, it is possible for a token sequence within $\enc(\worddict^*)$ to collide with a sequence from $\tokdict^* \setminus \enc(\worddict^*)$. 

To identify such collisions, we propose an algorithm $\getcoll$, described in \cref{alg:getcoll}. It takes as input a tokenizer $\tok$, a token sequence $\sigma$, and a depth parameter $d$. 
It iterates over all possible token sequences of length $d$ (given by the depth parameter) and checks if the decoded text matches text $\dec(\sigma)$. If the decoded text matches $\dec(\sigma)$, the token sequence is added to the collision set $Coll$. The function returns the collision set $Coll$ at the end of the iteration.

\begin{algorithm}[ht]
	\caption{$\getcoll(\tok, \sigma, d)$}
	\label{alg:getcoll}
	\begin{algorithmic}[1]    
		\State \textbf{Input}: Tokenizer $\tok$, token sequence $\sigma$, and depth parameter $d \geq 1$
		\State \textbf{Output}: The set of $d$ length token sequences that map to the same text as $\sigma$
		\State $Coll \gets \emptyset$
		\For{$\sigma' \in \tokdict^{d}$}
		\If{$\dec(\sigma') = \dec(\sigma)$}
		\State $Coll\gets Coll \cup \{\sigma'\}$
		\EndIf
		\EndFor
		\State \Return $Coll$
	\end{algorithmic}
\end{algorithm}

\end{document}